\documentclass[lettersize,journal]{IEEEtran}
\usepackage{amsmath,amsfonts}
\usepackage{algorithm}
\usepackage{algpseudocode}
\usepackage{amssymb}
\usepackage{array}
\usepackage{xcolor}
\usepackage[caption=false,font=normalsize,labelfont=rm,textfont=rm]{subfig}
\usepackage{textcomp}
\usepackage{stfloats}
\usepackage{url}
\usepackage{verbatim}
\usepackage{graphicx}
\usepackage{pifont}
\usepackage{booktabs}
\usepackage{cite}

\newcommand{\hollowstar}{\text{\ding{73}}}
\usepackage{multirow}
\newtheorem{principle}{Principle}

\begin{document}

\title{Causality-inspired Latent Feature Augmentation for Single Domain Generalization}

\author{Jian Xu, Chaojie Ji, Yankai Cao, Ye Li,~\IEEEmembership{Senior Member,~IEEE}, and Ruxin Wang,~\IEEEmembership{Member,~IEEE}

\thanks{This paper was supported in part by the National Key R\&D Program of China under Grant 2022YFA1008300, in part by the National Natural Science Foundation of China under Grant 62102410 and Grant 62206269, and in part by the Guangdong Provincial Science and Technology Plan under Grant 2022A1515011557 and Grant 2022B1515130009, and in part by the Shenzhen Science and Technology Program under Grant JCYJ20220818101216035. (Corresponding author: Ruxin Wang)} 
\thanks{Jian Xu, Ye Li and Ruxin Wang are with the Shenzhen Institute of Advanced Technology, Chinese Academy of Sciences, Shenzhen 518055, China. (e-mail: jianxu9603@gmail.com; ye.li@siat.ac.cn; rx.wang@siat.ac.cn)}
\thanks{Chaojie Ji is with the Department of Mathematics, The University of British Columbia, Vancouver, British Columbia, V6T 1Z2, Canada. (e-mail: chaojiej@math.ubc.ca)}
\thanks{Yankai Cao is with the Department of Chemical and Biological Engineering, The University of British Columbia, Vancouver, British Columbia, V6T 1Z2, Canada. (e-mail: yankai.cao@ubc.ca)}
}

\markboth{Journal of \LaTeX\ Class Files,~Vol.~14, No.~8, August~2021}%
{Shell \MakeLowercase{\textit{et al.}}: A Sample Article Using IEEEtran.cls for IEEE Journals}


\maketitle


\begin{abstract}

Single domain generalization (Single-DG) intends to develop a generalizable model with only one single training domain to perform well on other unknown target domains. Under the domain-hungry configuration, how to expand the coverage of source domain and find intrinsic causal features across different distributions is the key to enhancing the model’s generalization ability. Existing methods mainly depend on the meticulous design of finite image-level transformation techniques and learning invariant features across domains based on statistical correlation between samples and labels in source domain. This makes it difficult to capture stable semantics between source and target domains, which hinders the improvement of the model's generalization performance. 
In this paper, we propose a novel causality-inspired latent feature augmentation method for Single-DG by learning the meta-knowledge of feature-level transformation based on causal learning and interventions. Instead of strongly relying on the finite image-level transformation, with the learned meta-knowledge, we can generate diverse implicit feature-level transformations in latent space based on the consistency of causal features and diversity of non-causal features, which can better compensate for the domain-hungry defect and reduce the strong reliance on initial finite image-level transformations and capture more stable domain-invariant causal features for generalization. Extensive experiments on several open-access benchmarks demonstrate the outstanding performance of our model over other state-of-the-art single domain generalization and also multi-source domain generalization methods.

\end{abstract}  
\begin{IEEEkeywords}
Single domain generalization, causal intervention, latent feature augmentation.
\end{IEEEkeywords}

\maketitle
 
\section{Introduction}
\label{sec:intro}

Generally, the design of deep learning models relies on a strong distributional assumption that the training and testing samples are identically and independently distributed. However, due to the inevitable domain shift between the source (training) domain and the target (testing) domain in real-world scenarios, the model tends to experience a significant drop in generalization performance when tested on unseen scenarios directly~\cite{Ben-David_Blitzer_Crammer_Kulesza_Pereira_Vaughan_2010}.

\begin{figure}
  \centering
      {\includegraphics[width=0.95\linewidth]{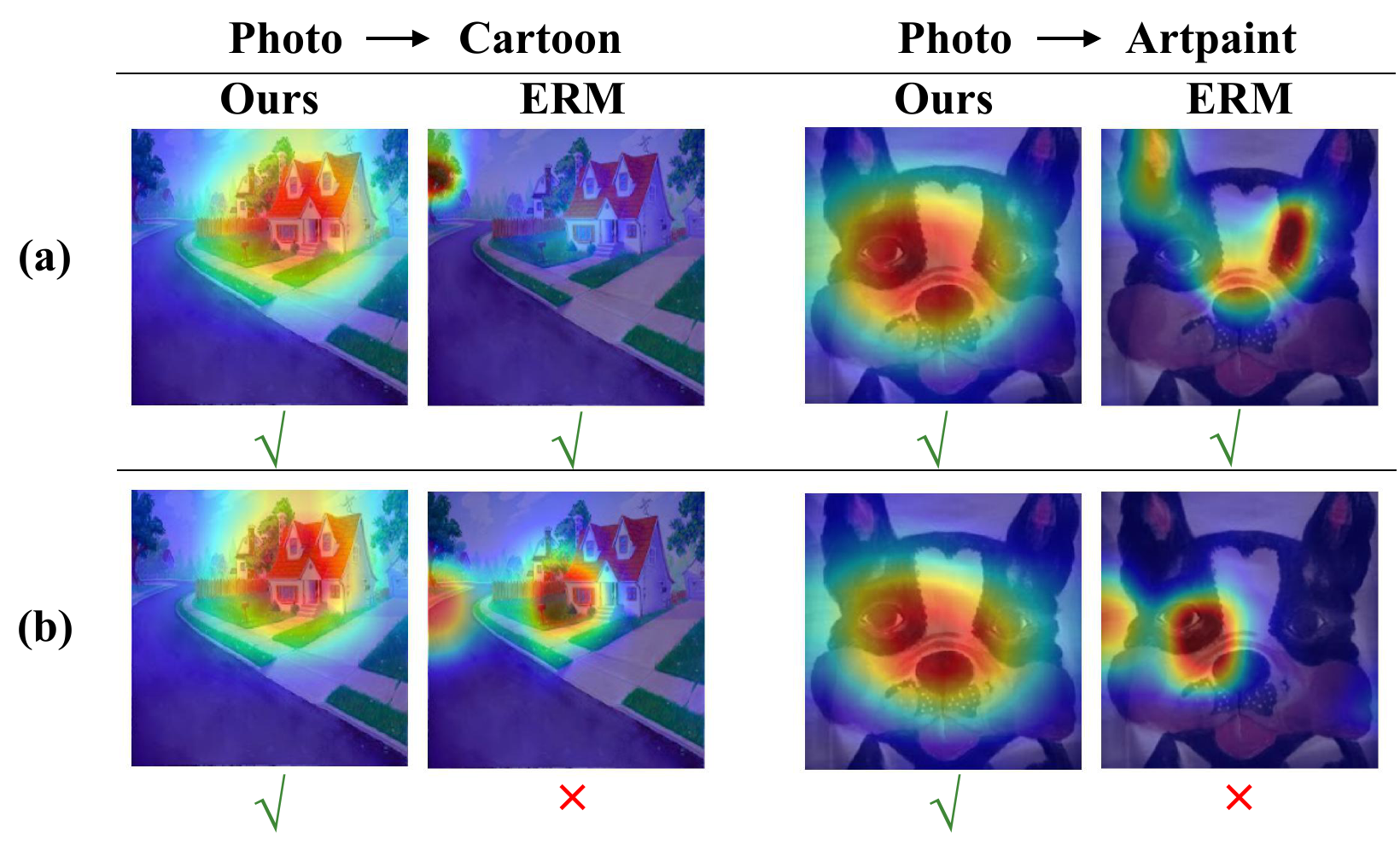} }
    \caption{Visual examples of ERM and Ours with (a) $16$ and (b) $5$ types of image-level transformation strategies on PACS dataset. ( $\color{green}{\checkmark}$: correct prediction, $\color{red}{\times}$: wrong prediction)}
    \label{fig1}
\end{figure}

Single domain generalization (Single-DG) aims to promote the cross-domain performance of the model trained on only a single domain, which has attracted increasingly more attention due to its practical application value \cite{Li_Gao_Cao_Huang_Weng_Mi_Yu_Li_Xia_2021}. 
Existing Single-DG works improve the model's generalization performance by data augmentation and learning domain-invariant representations \cite{Cugu_Mancini_Chen_Akata_2022, xu2023simde, Choi_Das_Choi_Yang_Park_Yun_2023, Wang_Luo_Qiu_Huang_Baktashmotlagh_2021, chen2023center,li2022uncertainty, Qiao_Zhao_Peng_2020, ballas2023cnn}. 
The data augmentation aims to generate diverse samples to expand the coverage of the source domain through various image-level transformation techniques for visual domain generalization tasks. However, the quantity of image-level transformations is finite and they need to be meticulously designed since they should not compromise the semantic information of the image itself. Therefore, although recent efforts have achieved better performance, the generalization ability of these models is strongly coupled with the way and quantity of finite image-level transformation.
Fig. \ref{fig1} shows a toy comparison. On the one hand, when the number of image-level transformation strategies is reduced from $16$ \footnote{$12$ photometric factors (Brightness, Contrast, Color, Sharpness, AutoContrast, Invert, Equalize, Solarize, SolarizeAdd, Posterize, NoiseSalt, NoiseGaussian) and $4$ geometric factors (ShearX, Shear-Y, Rotate, Flip) are mentioned in the latest Single-DG method~\cite{chen2023meta}.} to $5$\footnote{$4$ photometric factors (Brightness, Contrast, Color, Sharpness) and $1$ geometric factor (Rotate).}, ERM~\cite{koltchinskii2011oracle} exhibits the unstable phenomenon in discriminative feature learning and makes an incorrect identification due to the insufficient diversity of the source domain.
On the other hand, the obtained domain-invariant feature may establish spurious links between data and labels since the model does not explicitly distinguish causal and non-causal features among domains. The non-causal but discriminative feature in training data replaces the intrinsic semantic feature as the decision variable for prediction tasks, which results in poor generalization performance. 
For example, as illustrated in Fig. \ref{fig1}, with the number of transformation strategies fixed at $16$, although the model can make a correct prediction, it prefers the non-causal feature ``trees'' as the cues of the semantic class ``house'' within the images. This may be because trees as the shortcut features often appear around houses. 
More results and analysis will be discussed in Sec. \ref{analy_quan_way_img_level_trans} and \ref{example_visual_sub_sec}.

To resolve the above limits, we propose a causality-inspired learning model to extract latent features and disentangle causal factors and non-causal ones for Single-DG.
Firstly, we coin a causal structure graph representing the dependency among observed data, latent representation of data, causal variables, non-causal variables, and category labels. We propose that the augmentation on high-dimension latent space provides greater diversity compared with finite image-level transformation, which is also conducive to the exploration of causality.

Secondly, we introduce two meta-knowledge learning based on causality invariance to ensure the consistency of causal features and the diversity of non-causal features across domains. Specifically, for decoupled causal and non-causal factors, we design two encoders to learn two types of meta-knowledge about feature-level transformation in the latent space and generate augmented features by the generated diverse implicit feature-level transformations, which can reduce the strong reliance on the finite image-level transformation.

Lastly, we devise an effective intervention method based on the above initial/augmented causal variables and non-causal ones from latent feature space to obtain sufficient diversity of distributions and learn more stable causal features.

The contributions of this paper are summarized as:
\begin{itemize}
    \item We conduct a causal view on Single-DG problem and propose a novel latent feature augmentation paradigm based on causal learning and interventions, which is weakly coupled with finite initial image-level transformation strategies.
    
    \item At the feature level, we implement direct operations and interventions on causal/non-causal features by learning two types of meta-knowledge, which can generate more diverse latent feature spaces and capture more essential domain-invariant causal features for generalization.

    \item Extensive experiments are conducted on multiple single- and multi-source domain generalization benchmarks, giving rise to the state-of-the-art performances consistently. 
\end{itemize}

\section{Related Work}

\subsection{Single Domain Generalization}

Due to domain hungry, Single-DG is a more challenging and under-explored task in real scenarios compared with multi-source domain generalization (Multi-DG)~\cite{nguyen2023causal, 10163491} or domain adaptation (DA)~\cite{mansour2008domain}. 
Due to the lack of multiple source domains, most existing Single-DG methods primarily rely on data augmentation to expand the source domain distribution to enhance the model's generalization ability~\cite{zhao2020maximum, kim2023single, Qiao_Zhao_Peng_2020, Choi_Das_Choi_Yang_Park_Yun_2023, li2022uncertainty, chen2023center, zhang2023adversarial}. 
ACVC~\cite{Cugu_Mancini_Chen_Akata_2022} applied consistent visual attention across different augmentations of the same sample with visual corruptions to improve the generalization performance. 
SimDE~\cite{xu2023simde} learned diverse domain expansions by predicted entropy maximization and preserved semantics by cross-entropy minimization. 
PDEN~\cite{Li_Gao_Cao_Huang_Weng_Mi_Yu_Li_Xia_2021} progressively generated multiple domains to simulate various photometric and geometric transforms in unseen domains by joint learning two sub-networks.  
Wang et al.~\cite{Wang_Luo_Qiu_Huang_Baktashmotlagh_2021} proposed a style-complement module to create domain shifts between the generated images and the source images for enhancing the generalization power.
Chen et al.~\cite{chen2023center} proposed a center-aware adversarial augmentation technique to expand the source distribution.  
Although the above-mentioned methods have made considerable successes, the generalization ability is strongly coupled with the way and quantity of finite image-level transformation. Additionally, these models generate representation mainly based on statistical dependence without considering the underlying causal mechanisms.

\subsection{Causality for Domain Generalization}
Recently, some works introduced causality mechanisms to learn causal-invariant features and mitigate confoundings for domain generalization~\cite{lv2022causality, sheth2022domain, ouyang2022causality, chen2023meta}. Liu et al.~\cite{liu2021learning} proposed a causal semantic generative model based on causal reasoning to separate the semantic factors and variation factors. 
Lv et al.~\cite{lv2022causality} extracted the causal factors and reconstructed the invariant causal mechanisms from the inputs constructed from a mix of causal factors and non-causal factors. 
Ouyang et al.~\cite{ouyang2022causality} developed a causality-inspired data augmentation approach for medical image segmentation on single source domain generalization, which transforms images to have diverse appearances via randomly-weighted shallow convolutional networks and removes the confounders. Chen et al.~\cite{chen2023meta} presented a `simulate-analyse-reduce' learning paradigm and a meta-causal learning method to infer the causes of domain shift. The above methods are influenced by fixed image-level transformation strategies and their quantities, and they don't fully leverage the information from both the original data and augmented data, which limits the ability to expand unseen distributions and further affects the learning of domain-invariant causal features.

\section{Method}

\subsection{A Causal View on Single-DG}

\label{subsec:causal_view}

\begin{figure}
  \centering
      {\includegraphics[width=0.95\linewidth]{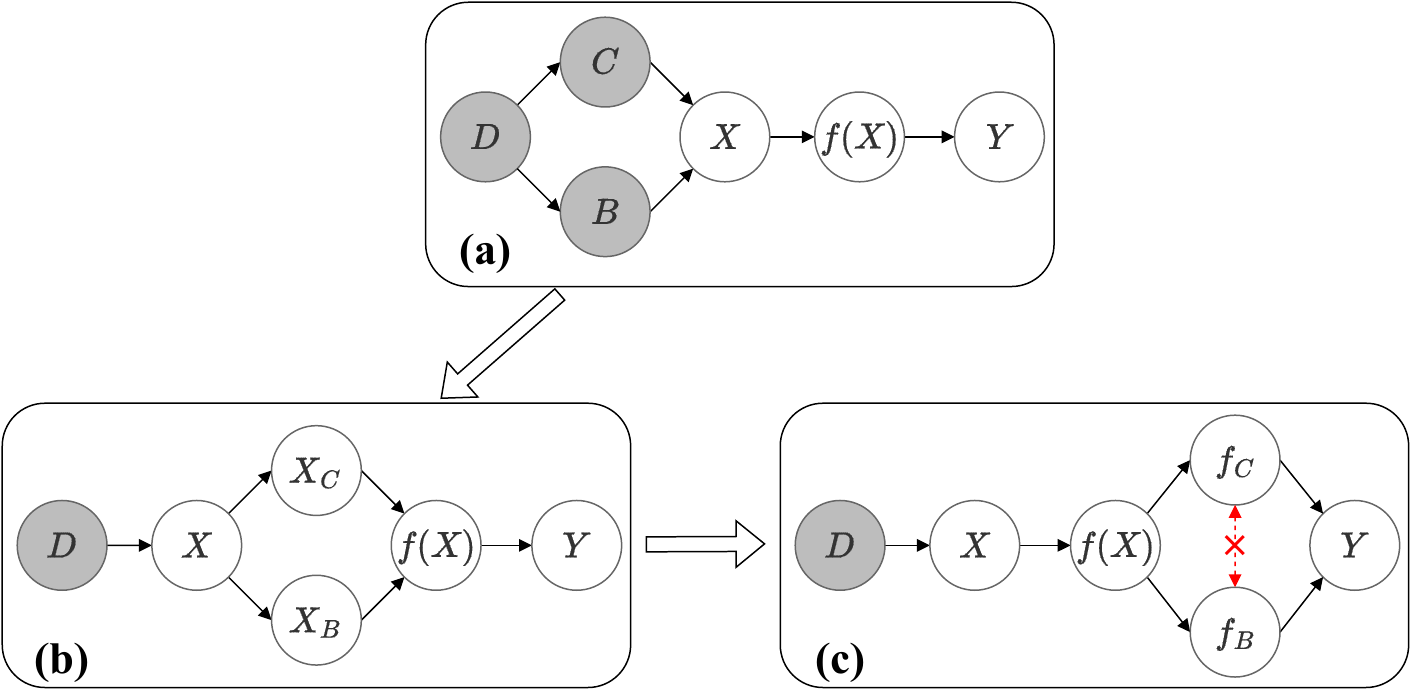} }
    \caption{Causal views on Single-DG. Grey variables represent unobservable variables in practice.}
    \label{causal_view}
\end{figure}

Here, we first model an initial causal dependency in Single-DG tasks with the assistance of the Structure Causal Model (SCM) as a preliminary. Then, we elaborate on how it evolves to our novel SCM focusing on feature level. 

The initial version of the SCM is depicted in Fig. \ref{causal_view}a which is built according to the following three observations:

\begin{itemize}
        \item $\boldsymbol{B\leftarrow D\rightarrow C}$: 
        Domain variable $D$ consists of two parts, i.e., causal (domain-invariant) component $C$ which faithfully reflects essential semantic attributes of data that is independent of domains, while non-causal (domain-specific) one $B$ embeds non-essential prior which relies on domains. They are all unobservable.
        \item $\boldsymbol{C\rightarrow X \leftarrow B}$: An observable image data $X$ is determined by the above two variables $C$ and $B$.
        \item $\boldsymbol{X \rightarrow f(X) \rightarrow Y}$: Variable $f(X)$ is the latent feature representation of the given data $X$ which is expected to be isolated to any change to a domain. $Y$ indicates the output category of $X$. 
\end{itemize}

Since the domain variable $D$ is always unobservable in the real world, an alternative approach to approximately capture it resorts to the technique of data augmentation, SCM of which is shown in Fig. \ref{causal_view}b. Instead of observing $D$, the interest of this variant SCM switches to $X$. Comparing Fig. \ref{causal_view}a and \ref{causal_view}b, the locations of variables $C$ and $B$ are moved forward.
Concretely, the variables $C$ and $B$ (determined by $D$ and acting on $X$) are replaced by variables $X_C$ and $X_B$ (currently affected by $X$ and acting on $f(X)$). The motivation here is that $X$ is an accessible variable which makes causal dependency turns to be observable. Nevertheless, the design incurs one problem. It is about the existence of backdoor path $X_C\leftarrow X \rightarrow X_B \rightarrow f(X) \rightarrow Y$. This backdoor path will cause $X_C$ to establish a spurious correlation with $Y$ due to the confounder $X_B$.

For the limitation, on top of image-level augmentation, some methods~\cite{ouyang2022causality, chen2023meta} introduce the technique of backdoor adjustment with the \emph{do}-calculus to cut off these backdoor paths, which can be formulated as:
\begin{equation} 
    \begin{aligned}
        P(Y|do(X_C)) = \sum_{x_b \in \mathcal{B}} P(Y|X_C, x_b)P(x_b), 
    \end{aligned}
    \label{cut_off_doXc}
\end{equation}
where $\mathcal{B}$ is the set of all confounders. Since enumerating all confounders $X_B=x_b, x_b \in \mathcal{B}$ is impractical, existent approaches apply a finite set of possible confounders to approximate the infinite set. 
\begin{equation}
    \begin{aligned}
       \sum_{\hat{x}_b \in \mathcal{\hat{B}}} P(Y|X_C, \hat{x}_b)P(\hat{x}_b) \approx \sum_{x_b \in \mathcal{B}} P(Y|X_C, x_b)P(x_b). 
    \end{aligned}
    \label{ori_data_augmentation}
\end{equation}
In Eq. (\ref{ori_data_augmentation}), $\mathcal{\hat{B}} = T(X_B)$ represents a series of finite elementary image transformations (e.g., rotation, dilation, and etc.), which is employed as the confounders set.
 
As the result, a ``good'' estimation of causal feature learning in Eq. (\ref{cut_off_doXc}) should follow two principle:

\begin{principle}
\label{princ_1}
Transformation should not compromise the semantic information  part. (The way of transformation) 
\end{principle}
\begin{principle}
\label{princ_2}
$|\mathcal{\hat{B}}|$ tends to  $|\mathcal{B}|$ where $|\cdot|$ represents the set size. The limited size of the transformation set results in unreliable estimates for causal feature learning. (The quantity of transformation)
\end{principle}
However, in real scenarios, image-level transformations are often finite and lack diversity, posing a challenge for domain-hungry Single-DG. Therefore, we propose a novel SCM focusing on feature level. Instead of purely performing the image-level transformation, we aim to further magnify the diversity of non-causal variable $X_B$, while, the causal variable and non-causal variable can be treated separately.
The proposed SCM is illustrated in Fig. \ref{causal_view}c. The main difference between ours and others is that we further defer the intervention, observation, and estimation of $X_C$ and $X_B$. Concretely, in our SCM, we disentangle variables $f_C$ and $f_B$ from observable variable $f(X)$ as high-dimension latent representation space is formed for $X$. The core benefit is that the amount of available transformation in high-dimension latent representation space is much larger than that of image transformation in a two dimension space. Secondly, we can transform causal variables and non-causal ones respectively, which leads to more precise estimation for them.
As a result, our method learns the interventional distribution $P(Y|do(f_C))$ as follows:
\begin{equation}
    \begin{aligned}
         P(Y|do(f_C))  \approx \sum_{\hat{f}_b \in \mathcal{\hat{B}}} P(Y|f_C, \hat{f}_b)P(\hat{f}_b),
    \end{aligned}
    \label{do_fc}
\end{equation}
where $\mathcal{{\hat B}}$ contains diverse transformations in latent space. In comparison to finite image-level transformations, we perform feature augmentation in latent space to generate more diverse implicit feature-level transformations based on the consistency of augmented semantic information, which can better follow \emph{Principle \ref{princ_1}} and \emph{\ref{princ_2}}, achieving more stable and reliable causal feature estimation as Eq. (\ref{do_fc}).

\subsection{Framework}

\begin{figure*}
  \centering
      {\includegraphics[width=0.98\linewidth]{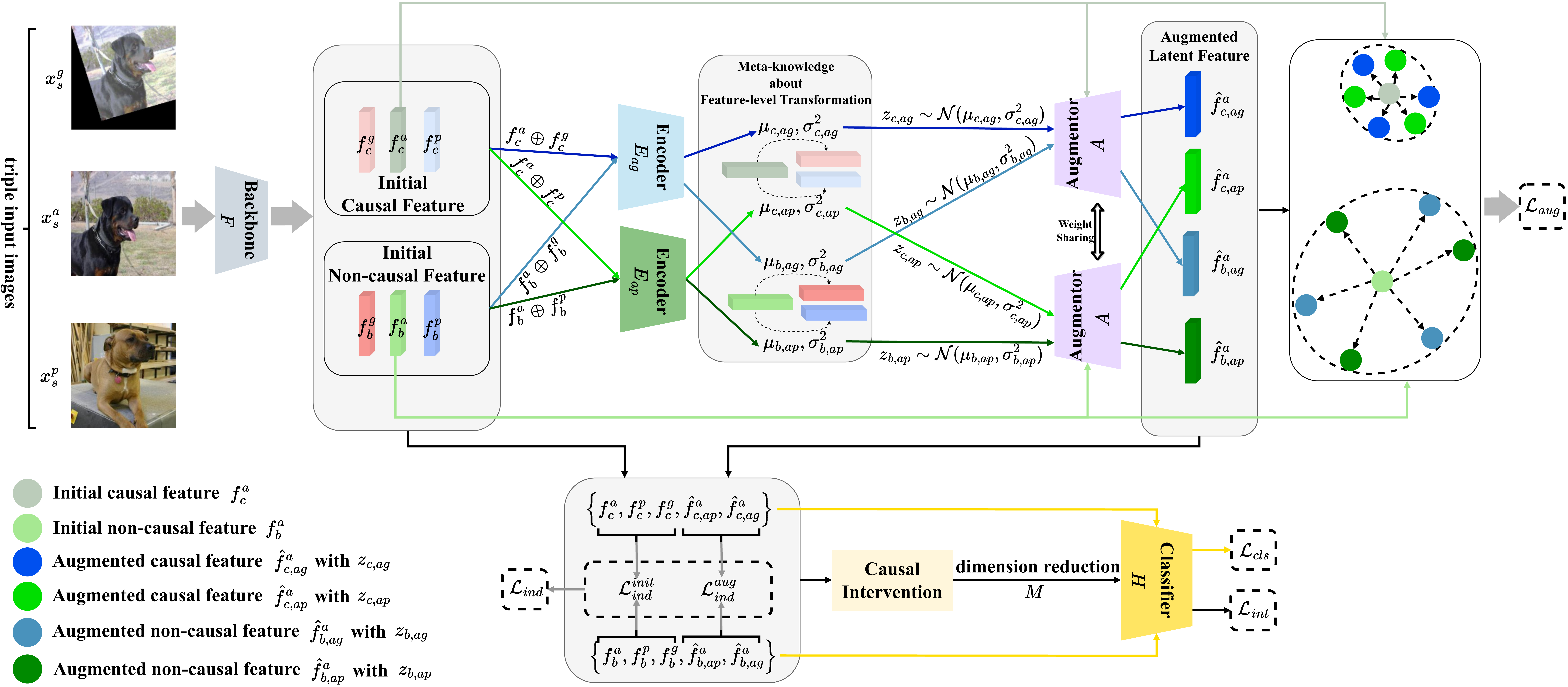} }
    \caption{Schematic of the proposed method. We generate mini-batches of triples in each iteration. Initial features extracted by $F$ is decomposed into causal and non-causal features. Two different ``Encoders'' learn two types of meta-knowledge about feature-level transformation. With the learned meta-knowledge, we generate  diverse implicit feature-level transformations by sampling. Then we perform latent feature augmentation by a shared ``Augmentor''. $\mathcal{L}_{ind}$ and $\mathcal{L}_{cls}$ encourage the independence and decoupling between latent causal and non-causal features. $\mathcal{L}_{aug}$ encourages the diversity of augmented latent non-causal features and the consistency of augmented latent causal features. $\mathcal{L}_{int}$ is the loss function about causal intervention.   
    }
    \label{framework}
\end{figure*}

\subsubsection{Causal and Non-causal Features Decoupling Representation}

To implement directed augmentation and intervention on causal/non-causal features based on two types of meta-knowledge learning, given a source domain $D_s=\{(x_s^i,y_s^i)\}_{i}$, from which we first generate mini-batches of triples in each training iteration. 
Each triple is in the form of $\{(x_s^a, y_s^a), (x_s^p, y_s^p), (x_s^g, y_s^g)\}$, where $x_s^a$ is an anchor sample, $x_s^p$ is a sample from $D_s$ with the same class of $x_s^a$, $x_s^g = T(x_s^a)$ represents that an augmented sample by the predefined explicit image-level transformation strategies $T$ on $x_s^a$, and labels $y_s^a=y_s^p=y_s^g$. For illustrative purposes, we define $v\in \{a,p,g\}$. 

For simplicity, we construct a feature extractor $F$ to extract the initial latent feature $f\in \mathbb{R}^d$ and explicit decompose it into two parts: causal features $f_c\in \mathbb{R}^{d/2}$ and non-causal features $f_b \in \mathbb{R}^{d/2}$, based on which we can obtain:
\begin{equation}
    \begin{aligned}
        &f_c^v, f_b^v = F(x_s^v). 
    \end{aligned}
    \label{decompose_feat}
\end{equation} 
Since the causal features are related to the ground-truth label while non-causal ones encode other trivial patterns or bias, we utilize a classifier $H$ to induce $f_c$ and $f_b$ to satisfy the properties of causal feature and non-causal feature respectively: 
\begin{equation}
\begin{aligned}
     \mathcal{L}_{cls}^{init} =\min_{F, H} \sum_{v\in \{a,p,g\}} \big( &-{y_s^v} \log (H(f_c^v)) \\
   &+ KL (y_{uniform}, H(f_b^v)) \big ),
\end{aligned}
\label{cb_cls_loss}
\end{equation}
where $KL(\cdot)$ indicates the Kullback-Leibler divergence, which is employed to align the distribution of non-causal factors with a uniform distribution $y_{uniform}$. Kindly noted that previous studies~\cite{NEURIPS2019_4c5bcfec,lin2021generative} have demonstrated a higher mutual information between the causal factor and label compared to that between the full images and label, due to the prevalent trivial patterns or noise. Consequently, the proposed decoupling strategy can prevent the converging of the causal features to plain full features. Furthermore, to enforce independence and decoupling between causal and non-causal features, we introduce a constraint term based on correlation defined as:
\begin{equation}
    \mathcal{L}_{ind}^{init} = \min_{F} \sum_v \frac{1}{2} ||\mathbf{C}_v-\mathbf{0}||_2^2,
    \label{ind_loss}
\end{equation}
where 
\begin{equation}
    \mathbf{C}_{v}=\frac{<f_c^v, f_b^v>}{||f_c^v||\;||f_b^v||}
\end{equation}
measures the correlation between $f_c^v$ and  $f_b^v$. 

\subsubsection{Latent Feature Augmentation}
\label{sub_sub_sec:data aug module}

To expand the latent feature space on the source domain and learn domain-invariant causal features, specifically, we learn two types of meta-knowledge about feature-level transformation based on the information between $x_s^a$ and $x_s^g$, as well as between $x_s^a$ and $x_s^p$ respectively. The former is the transformation between the original instance and its augmented instance with predefined image-level transformation strategies, while the latter is that between the original instance and the instance with an identical category.
We assume that the meta-knowledge represents nonlinear deformations that can augment the diversity of latent features, which are treated as the multivariate Gaussian distribution. With the meta-knowledge, we can generate more diversified implicit feature-level strategies, beyond being confined to predefined explicit image-level transformation strategies, thereby expanding the source distribution so that we can estimate Eq. (\ref{do_fc})  more accurately. 
Concretely, we define two encoders $E_{ag}$ and $E_{ap}$ to learn the meta-knowledge of nonlinear deformations from $f_c^a$ to $f_c^g$ and from $f_c^a$ to $f_c^p$. We then employ a shared Augmentor $A$ to augment the latent features based on the proposed two types of meta-knowledge. 

According to the theoretical analysis in Sec. \ref{subsec:causal_view}, we augment causal features $f_c$ and non-causal features $f_b$ separately, ensuring both the consistency of augmented causal features and the diversity of augmented non-causal features to learn better domain-invariant causal features. Given $t\in \{c,b\}$, $f_t^a$ and $f_t^g$ are concatenated and input to $E_{ag}$, while $f_t^a$ and $f_t^p$ are concatenated and input to $E_{ap}$. The output is represented as meta-knowledge about feature-level transformation, denoted as ${\mu_{t,ag}, \sigma^2_{t, ag}}$ for the first type of meta-knowledge and ${\mu_{t, ap}, \sigma^2_{t, ap}}$ for the second one respectively:
\begin{equation}
    \begin{aligned}
    & \mu_{t,ag}, \sigma^2_{t, ag} = E_{ag} (f_t^a\oplus f_t^g), \\
    & \mu_{t,ap}, \sigma^2_{t, ap} = E_{ap} (f_t^a\oplus f_t^p),
    \end{aligned}
    \label{meta_knowledge}
\end{equation}
where $\oplus$ represents the feature concatenation. By the meta-knowledge, we generate the implicit feature-level transformation strategies $z$ through sampling. Specifically, we use the reparameterization trick~\cite{kingma2013auto} for backpropagation computation:
\begin{equation}
    \begin{aligned}
        &z_{t,ag}=\mu_{t,ag} + \epsilon \Sigma_{t,ag}, \\
        &z_{t,ap}=\mu_{t,ap} + \epsilon \Sigma_{t,ap},
    \end{aligned}
    \label{z_ag_ai}
\end{equation}
where $\epsilon \sim \mathcal{N}(0,1)$ and $\Sigma=diag(\sigma)$.  Then, 
 the initial feature $f_t^a$ and the randomly sampled implicit feature-level transformation strategy $z$ are fed into Augmentor $A$ to generate two types of augmented latent features:
 \begin{equation}
    \begin{aligned}
        \hat{f}_{t,ag}^a = A(f_t^a\oplus z_{t,ag}), \\ 
        \hat{f}_{t,ap}^a = A(f_t^a\oplus z_{t,ap}).
    \end{aligned}
    \label{augment_latent_feature}
\end{equation}

Through the sampling operations as Eq. (\ref{z_ag_ai}) for $\lambda$ times, we perform latent feature augmentation for $\lambda$ times. As the iterations progress, each sampling  produces distinct implicit feature-level transformations, enabling the satisfaction of Principle \ref{princ_2}. This encourages the model to better estimate causal features as in Eq. (\ref{do_fc}).  Similar to Eq. (\ref{cb_cls_loss}) and Eq. (\ref{ind_loss}), we also enforce the decoupling and independence between the augmented latent causal and non-causal features, which is denoted as $\mathcal{L}_{cls}^{aug}$ and $\mathcal{L}_{ind}^{aug}$. Therefore, the overall losses for the constraint of decoupling and independence between causal and non-causal features are denoted as $\mathcal{L}_{cls}=\mathcal{L}_{cls}^{init} + \mathcal{L}_{cls}^{aug}$ and $\mathcal{L}_{ind} = \mathcal{L}_{ind}^{init} + \mathcal{L}_{ind}^{aug}$, respectively.  

Furthermore, we use Euclidean distance to measure the distance between initial latent features and augmented latent features:
\begin{equation}
    \begin{aligned}
        &d(f_t^a, \hat{f}_{t,ag}^a) = ||f_t^a - \hat{f}_{t,ag}^a||_2^2, \\
        &d(f_t^a, \hat{f}_{t,ap}^a) = ||f_t^a - \hat{f}_{t,ap}^a||_2^2.
    \end{aligned}
\end{equation}
Then we impose the constraints simultaneously in the causal feature space and the non-causal feature space jointly, augmenting the diversity of non-causal features under the condition ensuring the consistency of causal feature augmentation:
\begin{equation}
    \begin{aligned}
        \mathcal{L}_{aug}&= \sum_{v\neq a} \Big\{ \min_{F,E_{av},A} d(f_c^a, \hat{f}_{c,av}^a)  \\ 
         & + \max_{F,E_{av},A} \big\{ d(f_c^a, \hat{f}_{c,av}^a)-d(f_b^a, \hat{f}_{b,av}^a) + \delta, 0 \big\} \Big\},
    \end{aligned}
    \label{A_loss}
\end{equation}
where $\delta > 0$ is a small margin variable. 
In Eq. (\ref{A_loss}), the first term penalizes the consistency of augmented causal features which can follow Principle \ref{princ_1}, while the second term penalizes the diversity of augmented non-causal features. The first term is introduced to avoid increasing both $d(f_c^a, \hat{f}_{c,av}^a)$ and $d(f_b^a, \hat{f}_{b,av}^a)$ simultaneously, which contradicts the consistency constraint assumption.  Besides, on the one hand, $\delta$ prevents the situation where $d(f_c^a, \hat{f}_{c,av}^a)-d(f_b^a, \hat{f}_{b,av}^a)$ is always less than zero and never penalized. On the other hand, $\delta$ controls the intensity of the non-causal feature augmentation.

\subsection{Causal Intervention}
Furthermore, we estimate the Eq. (\ref{do_fc}) by making the causal intervention to obtain sufficient diversity of distributions and learn better domain-invariant causal features. Specifically, we define a ``Dimension Reduction'' linear layer $M$ in conjunction with the classifier $H$ to classify the ``intervened features'' combined by the fixed invariant causal features and diversely augmented variant non-causal features, and formulate the following loss function:
\begin{equation}
    \begin{aligned}
        \mathcal{L}_{int} = \min_{\Omega} 
        -\frac{1}{|\mathcal{\hat{C}}|\; |\mathcal{\hat{B}}|} 
         & \sum_{\tilde{f}_c\in \mathcal{\hat{C}} } \sum_{ \tilde{f}_b\in \mathcal{\hat{B}}} 
        \Big ( y_s^v \log H(M(\tilde{f}_c\oplus \tilde{f}_b))\\
        & + KL(H(\tilde{f}_c), H(M(\tilde{f}_c\oplus \tilde{f}_b)) \Big ),
    \end{aligned}
    \label{final_cls_loss}
\end{equation}
where $\mathcal{\hat{C}}=\{f_c^a, \hat{f}_{c,ap}^a, \hat{f}_{c,ag}^a, f_c^g , f_c^p \}$ and $\mathcal{\hat{B}}=\{f_b^a, \hat{f}_{b,ap}^a, \hat{f}_{b,ag}^a, f_b^g, f_b^p \}$ and $\Omega=\{F,E_{ap},E_{ag}, A, M, H \}$. 
In Eq. (\ref{final_cls_loss}), we treat all the initial and augmented non-causal features as confounders and randomly shuffle them. Then we make the intervention by $\tilde{f}_c\oplus \tilde{f}_b$. 
The first item of Eq. (\ref{final_cls_loss}) facilitates the correct classification of the intervened features, while the second item ensures consistency in the classification distribution between the intervened features and causal features.
Totally, Eq. (\ref{final_cls_loss}) makes our model to better estimate Eq. (\ref{do_fc}) and learn better causal features. Finally, the overall loss function is defined as:
\begin{equation}
    \mathcal{L} =  \mathcal{L}_{cls} + \alpha_1\mathcal{L}_{ind} + \alpha_2\mathcal{L}_{aug} + \alpha_3\mathcal{L}_{int}, 
    \label{total_loss}
\end{equation}
where the sensitivity analysis of three weighting coefficients $\alpha_1$, $\alpha_2$ and $\alpha_3$ is shown in Sec. \ref{sens_loss_coeffi_subsec}. 

\section{Experiments}

\subsection{Datasets}

\textbf{PACS}~\cite{li2017deeper} consists of 4 domains: art painting, cartoon, photo, and sketch. There are $9991$ images of $7$ categories. We use one domain as the source domain, and the rest three domains as the target domains. \textbf{Digits} consists of $5$ datasets: MNIST~\cite{Lecun_Bottou_Bengio_Haffner_1998}, MNIST-M~\cite{ganin2015unsupervised}, SVHN~\cite{netzer2011reading}, USPS~\cite{Hull_1994}, and SYN~\cite{ganin2015unsupervised}, with 10 categories. We use MNIST as the source domain, and the other four datasets as the target domains. Following~\cite{chen2023meta}, the first $10000$ images in MNIST are used for training. \textbf{CIFAR10-C}~\cite{hendrycks2019benchmarking} is corrupted from the test set of the CIFAR10 dataset~\cite{krizhevsky2009learning} by $19$ corruption types with 5 levels of severity and 10 categories totally. A higher level means more serious corruption. We use CIFAR10 as the source domain, and images with different severity levels in CIFAR10-C form various target domains. 
Additionally, we introduce two additional datasets, \textbf{Office-Home}~\cite{venkateswara2017deep} ($15,588$ images, $65$ classes, and $4$ domains) and \textbf{DomainNet}~\cite{peng2019moment} ($586,575$ images, $345$ classes, and $6$ domains), to extend and validate our method in multi-source domain generalization (Multi-DG) setting. 

\subsection{Implementation Details}

We implement our framework by employing the same $16$ initial image-level transformation strategies following~\cite{chen2023meta}. 
For a fair comparison, following common practices~\cite{chen2023meta, Wang_Luo_Qiu_Huang_Baktashmotlagh_2021,Li_Gao_Cao_Huang_Weng_Mi_Yu_Li_Xia_2021,Qiao_Zhao_Peng_2020}, we use the same backbone as other Single-DG methods. Specifically, for Digits dataset, we use ConvNet~\cite{lecun1989backpropagation} as the backbone. For CIFAR10 dataset, the WRN~\cite{zagoruyko2016wide} with $16$ layers and widen factor $4$ is employed as the backbone. For PACS dataset, we use ResNet18~\cite{he2016deep} pretrained on ImageNet as the backbone. 

For the ``Encoder'' and ``Augmentor'' in the proposed framework, we employ the same architecture with different parameters: two fully-connected layers with ReLU following each layer.
(``Encoder'': $1024\rightarrow512$ for PACS dataset, $1024\rightarrow128$ for CIFAR10 dataset, and $1024\rightarrow64$ for Digits dataset. ``Augmentor'': $1024\rightarrow1024$ for PACS dataset, $1024\rightarrow128$ for CIFAR10 dataset, and $1024\rightarrow512$ for Digits dataset). 

In the training process of our model, the mini-batch of triples in each iteration is formed by randomly sampling from the dataset and keeping the class balanced. Specifically, for each category, we form $4$ triples for PACS and $8$ triples for both Digits and CIFAR10. The maximum training iterations for all datasets are set to $40000$. We employ Adam as the optimizer of our model. In practice, for PACS dataset, the initial learning rate is set to $10^{-4}$. For Digits and CIFAR10 datasets, the initial learning rates are set to $10^{-3}$. During the training procedure, the learning rate is annealed by half for every $10000$ iterations. The margin $\delta$ in Eq. (\ref{A_loss}) is set to $2$.

The training process of the proposed model is shown in \textbf{Algorithm \ref{training_al}}. We repeat the experiment ten times with different random seeds and report the average results.

\begin{algorithm}[!t]
\caption{Training process of our proposed model} 
\begin{algorithmic}[1]
    \Require The source domain $D_s$, the pre-defined explicit image-level  transformation strategy set $\mathcal{T}$.
    \Ensure $\Omega$ = \{$F$, $E_{ag}$, $E_{ap}$, $A$, $M$, $H$\}.
    \While{not converge}
        \State{Randomly sample image-level transformation strategies $T$ from $\mathcal{T}$.}
        \State{Sample $x_s^a$ and $x_s^p$ from $D_s$ and generate $x_s^g$ by $T(x_s^a)$ to generate mini-batch of triples.}
        \State{Get initial causal and non-causal features $ \mathcal{F}_{c} =\{f_c^a,f_c^p,f_c^g\}$, $\mathcal{F}_{b} =\{f_b^a,f_b^p,f_b^g\}$.}
        \State{$t\in \{c,b\}$.}
        \State{Get two types of meta-knowledge $\mu_{t,ag},\sigma^2_{t,ag}$ and $\mu_{t,ap},\sigma^2_{t,ap}$ about feature-level transformation as Eq. (\ref{meta_knowledge}).}  
        \For{1, 2, $\cdots$, $\lambda$}
            \State{ 
            Sample implicit feature-level transformation $z_{t,ag}$ and $z_{t,ap}$ as Eq. (\ref{z_ag_ai}). }
            \State{Get augmented features $\hat{f}_{t,ag}^a$ and $\hat{f}_{t,ap}^a$ as Eq. (\ref{augment_latent_feature}). }
            \State{$\mathcal{F}_c := \mathcal{F}_c \cup \{\hat{f}_{c,ag}^a, \hat{f}_{c,ap}^a\}$.}
            \State{$\mathcal{F}_b := \mathcal{F}_b \cup \{\hat{f}_{b,ag}^a, \hat{f}_{b,ap}^a\}$.}
        \EndFor
        \State{Update $\Omega$ by Eq. (\ref{total_loss}). }
    \EndWhile
\end{algorithmic}
\label{training_al}
\end{algorithm}

\subsection{Performance Comparisons}

We extensively compare our approach against existing well acknowledged Single-DG methods, which can
be classified into four groups, including the baseline (ERM~\cite{koltchinskii2011oracle}), domain-invariant features learning (CCSA~\cite{motiian2017unified}, d-SNE~\cite{xu2019d}, JiGen~\cite{carlucci2019domain}), data augmentation (ADA~\cite{volpi2018generalizing}, M-ADA~\cite{Qiao_Zhao_Peng_2020}, ME-ADA~\cite{zhao2020maximum}, L2D~\cite{Wang_Luo_Qiu_Huang_Baktashmotlagh_2021}, PDEN~\cite{Li_Gao_Cao_Huang_Weng_Mi_Yu_Li_Xia_2021}, SimDE~\cite{xu2023simde}, AA~\cite{cubuk2019autoaugment}, RA~\cite{cubuk2020randaugment}, CADA~\cite{chen2023center}, RSDA~\cite{volpi2019addressing}, Pro-RandConv~\cite{Choi_Das_Choi_Yang_Park_Yun_2023}), and causality with data augmentation (MCL~\cite{chen2023meta}). 

The comparison results on PACS, Digits, and CIFAR10-C are shown in Table \ref{PACS_exp}, Table \ref{DIGITS_exp}, and Table \ref{CIFAR10C_exp} respectively. Overall, our proposed method achieves the best average classification accuracy compared to other methods on all datasets, which shows the effectiveness of our method. Compared to the methods of using image-level transformation, our method also outperforms them, which indicates that our method can generate richer domain changes and align better with the target domain. 
Our method outperforms the domain-invariant representation based methods CCSA~\cite{motiian2017unified}, d-SNE~\cite{xu2019d}, and JiGen~\cite{carlucci2019domain} by a large margin. This demonstrates the importance of excavating the intrinsic causal feature across domains, in place of the superficial statistical correlation. 
Besides, we also compared our approach with the latest method MCL~\cite{chen2023meta} based on causality learning, our method gets better performance on all datasets, which indicates that our method can learn more stable domain-invariant causal features by latent feature augmentation. In particular, our method gets significant generalization performance improvement on more challenging and difficult tasks with larger domain shift (such as Photo on PACS, SVHN on Digits, level5 on CIFAR10-C with low-resolution images) profit by the proposed causal learning and intervention strategies in the feature space.

\begin{table}[!t]
  \centering
  \small
  \caption{Comparison with state-of-the-art Single-DG methods on PACS benchmark. (Average accuracy(\%)) }
  \begin{tabular}{@{}l|cccc|c@{}}
    \toprule
    Method  &Artpaint & Cartoon & Photo & Sketch & Avg. \\
    \midrule
    ERM~\cite{koltchinskii2011oracle} & 70.90 & 76.50  & 42.20 & 53.10 & 60.70 \\ \hline
    ADA~\cite{volpi2018generalizing} & 71.56 & 76.84  & 43.66 & 52.36 & 61.11 \\
    ME-ADA~\cite{zhao2020maximum} & 71.52 & 76.83  & 46.32 & 46.22 & 60.22 \\
    RSC~\cite{huang2020self} & 73.40 & 75.90 & 41.60 & 56.20 & 61.80\\
    RSC+ASR~\cite{fan2021adversarially} & 76.70 & 79.30 & 54.60 & 61.60 &68.10\\ 
    L2D~\cite{Wang_Luo_Qiu_Huang_Baktashmotlagh_2021} & 77.08 &75.21 &54.14 &55.21 &65.41 \\ 
    PDEN~\cite{Li_Gao_Cao_Huang_Weng_Mi_Yu_Li_Xia_2021} & 76.43 &73.87 &58.52 &53.92 &65.68 \\ 
    SimDE~\cite{xu2023simde} & 78.52 &76.14 &59.32 &56.39 &67.59 \\ 
    Pro-RandConv~\cite{Choi_Das_Choi_Yang_Park_Yun_2023} & 76.98 &78.54 &62.89 &57.11 &68.88 \\
    FACT-DE~\cite{xu2023fourier}&75.70 &78.40 & 50.70 &62.80 &66.90 \\ 
    ALT~\cite{gokhale2023improving}&75.70 &77.30 & 55.10 &50.70 &64.70 \\ 
     CADA~\cite{chen2023center} & 76.33 &79.08 &56.65 &61.59 &68.41 \\ 
    MCL~\cite{chen2023meta} & 77.13 &80.14 &59.60 &62.55 &69.86 \\
    \hline
    Ours  & \textbf{81.26} &\textbf{81.63} &\textbf{63.68} &\textbf{63.44} &\textbf{72.50} \\
    \bottomrule
  \end{tabular}
  \label{PACS_exp}
\end{table}

\begin{table}[!t]
  \centering
  \small
  \caption{Comparison with state-of-the-art Single-DG methods on Digits benchmark. (Average accuracy(\%)) }
  \begin{tabular}{@{}l|cccc|c@{}}
    \toprule
     Method  &SVHN & SYN & MNIST-M & USPS & Avg. \\
    \midrule
    ERM~\cite{koltchinskii2011oracle} & 27.83 & 39.65  & 52.72 & 76.94 & 49.29 \\ \hline
    CCSA~\cite{motiian2017unified} & 25.89 & 37.31  & 49.29 & 83.72 & 49.05 \\
    d-SNE~\cite{xu2019d} & 26.22 & 37.83  & 50.98 & 93.16 & 52.05 \\
    JiGen~\cite{carlucci2019domain} & 33.80 & 43.79  & 57.80 & 77.15 & 53.14 \\ 
    ADA~\cite{volpi2018generalizing} & 35.51 & 45.32  & 60.41 & 77.26 & 54.62 \\
    M-ADA~\cite{Qiao_Zhao_Peng_2020} & 42.55 & 48.95  & 67.94 & 78.53 & 59.49 \\
    ME-ADA~\cite{zhao2020maximum} & 42.56 & 50.39  & 63.27 & 81.04 & 59.32 \\
    PDEN~\cite{Li_Gao_Cao_Huang_Weng_Mi_Yu_Li_Xia_2021} & 62.21 & 69.39  & 82.20 & 85.26 & 74.77 \\
    L2D~\cite{Wang_Luo_Qiu_Huang_Baktashmotlagh_2021} & 62.86 & 63.72  & 87.30 & 83.97 & 74.46 \\
    AA~\cite{cubuk2019autoaugment} & 45.23 & 64.52  & 60.53 & 80.62 & 62.72 \\
    RA~\cite{cubuk2020randaugment} & 54.77 & 59.60  & 74.05 & 77.33 & 66.44 \\
    RSDA~\cite{volpi2019addressing} & 47.40 & 62.00  & 81.50 & 83.10 & 68.50 \\
    RSDA+ASR~\cite{fan2021adversarially} & 52.80 & 64.50  & 80.80 & 82.40 & 70.10 \\
    MetaCNN~\cite{wan2022meta} & 66.50 & 70.66  & \textbf{88.27} & 89.64 & 78.76 \\
    SimDE~\cite{xu2023simde} & 66.08 & 70.04  & 84.90 & 86.56 & 76.89  \\
    Pro-RandConv~\cite{Choi_Das_Choi_Yang_Park_Yun_2023} & 69.67 & 79.77  & 82.30 & 93.67 & 81.35 \\ 
    MCL~\cite{chen2023meta} & 69.94 & 78.47  & 78.34 & 88.54 & 78.82 \\
    \hline
    Ours & \textbf{70.10} & \textbf{79.85}  & 81.65 & \textbf{94.28} & \textbf{81.47} \\
    \bottomrule
  \end{tabular}
  \label{DIGITS_exp}
\end{table}

\begin{table}[!t]
  \centering
  \small
  \caption{Comparison with state-of-the-art Single-DG methods on CIFAR10-C benchmark. (Average accuracy(\%))}
  \begin{tabular}{@{}l|ccccc|c@{}}
    \toprule
    Method  &level1 &level2 & level3 & level4 & level5 & Avg. \\
    \midrule
    ERM~\cite{koltchinskii2011oracle}  & 87.80 & 81.50  & 75.50 & 68.20 & 56.10 & 73.82 \\ \hline
    ADA~\cite{volpi2018generalizing} & 88.30 & 83.50  & 77.60 & 70.60 & 58.30 & 75.66 \\
    M-ADA~\cite{Qiao_Zhao_Peng_2020} & 90.50 & 86.80  & 82.50 & 76.40 & 65.60 & 80.36 \\
    PDEN~\cite{Li_Gao_Cao_Huang_Weng_Mi_Yu_Li_Xia_2021} & 90.62 & 88.91  & 87.03 & 83.71 & 77.47 & 85.55 \\
    AA~\cite{cubuk2019autoaugment} & 91.42 & 87.88  & 84.10 & 78.46 & 71.13 & 82.60\\
    RA~\cite{cubuk2020randaugment} & 91.74 & 88.89 & 85.82 & 81.03 & 74.93 & 84.48 \\ 
    MCL~\cite{chen2023meta} & 92.38 & 91.22  & 89.88 & 87.73 & 84.52 & 89.15 \\
    \hline
    Ours  & \textbf{92.81} &\textbf{91.92} &\textbf{90.11} &\textbf{88.57} &\textbf{86.60} &\textbf{90.00} \\
    \bottomrule
  \end{tabular}
  \label{CIFAR10C_exp}
\end{table}

\subsection{Effect of Causal Learning}

We conduct an ablation study on different constraints of our model, which is shown in Table \ref{Ablation_model_causality}. The ``baseline'' is the model that utilizes the standard cross entropy loss to optimize. With image-level transformation strategies, the ``+$T$'' performs better than the baseline model, which shows that the image-level transformation can expand the source domain. In our work, based on initial image-level transformation, we simulate domain shift by learning two types of meta-knowledge in the latent feature space with causal learning and intervention mechanism. Below, on the basis of the ``baseline+$T$'', we specifically examine the impact of different  constraints on our model. As we can see, ``$+\mathcal{L}_{cls}$'' and  ``$+\mathcal{L}_{cls}+\mathcal{L}_{ind}$'' both perform better than the ``baseline+$T$''. The ``$\mathcal{L}_{cls}$'' decouples the entire latent feature into causal and non-causal features by means of a shared classifier, while ``$\mathcal{L}_{ind}$'' promotes the independence of causal and non-causal features. The combination of these two constraints encourages the model to learn more robust domain-invariant causal features, thus improving the generalization capability. Additionally, $\mathcal{L}_{aug}$ can lead to a higher improvement, which enhances the cross-domain invariance of causal features and diversity of non-causal features. Remarkably, after introducing the causal intervention constraints $\mathcal{L}_{int}$, the generalization performance is further promoted in that the causal intervention can enable the model to estimate the domain-invariant causal features better as Eq. (\ref{do_fc}) and learn more robust causal features. 
\begin{figure*}[!t]
  \centering
     \subfloat[baseline \\ (acc=42.20)]{\includegraphics[width=0.199\linewidth]{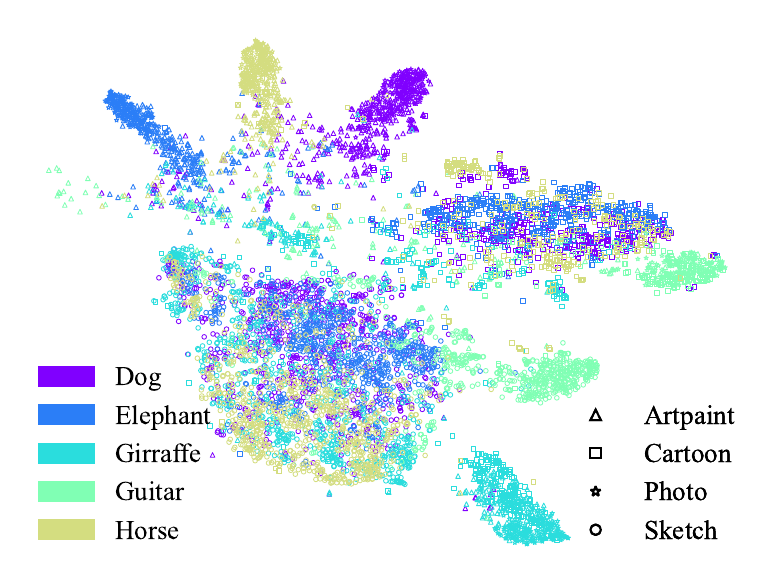} \label{fig:subfig1} }  
      \subfloat[baseline+T \\ (acc=56.81)]{\includegraphics[width=0.199\linewidth]{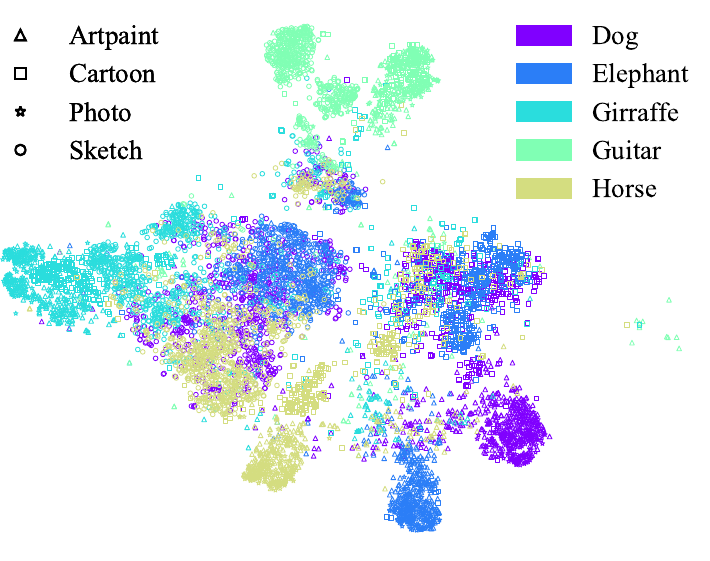} \label{fig:subfig2} }  
       \subfloat[+$\mathcal{L}_{cls}$+$\mathcal{L}_{ind}$ \\ (acc=60.05)]{\includegraphics[width=0.199\linewidth]{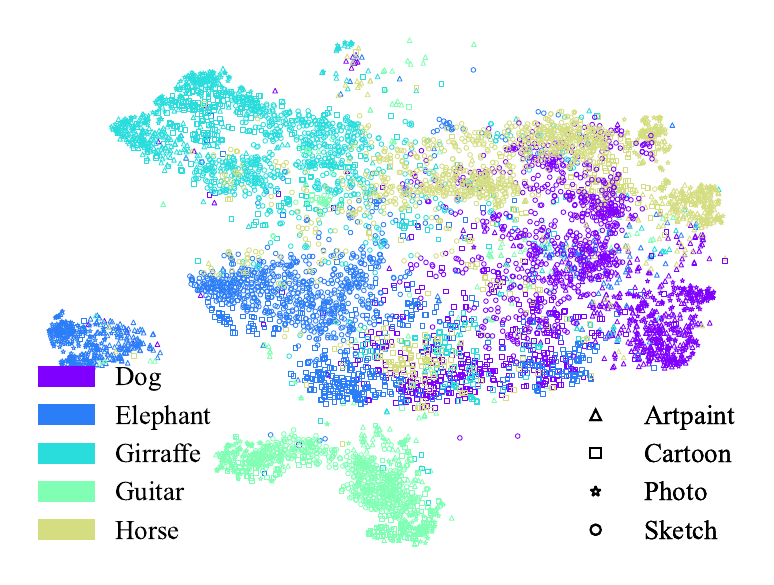} \label{fig:subfig3} } 
        \subfloat[+$\mathcal{L}_{cls}$+$\mathcal{L}_{ind}$+$\mathcal{L}_{aug}$  \\ \centering{(acc=62.74)}]{\includegraphics[width=0.199\linewidth]{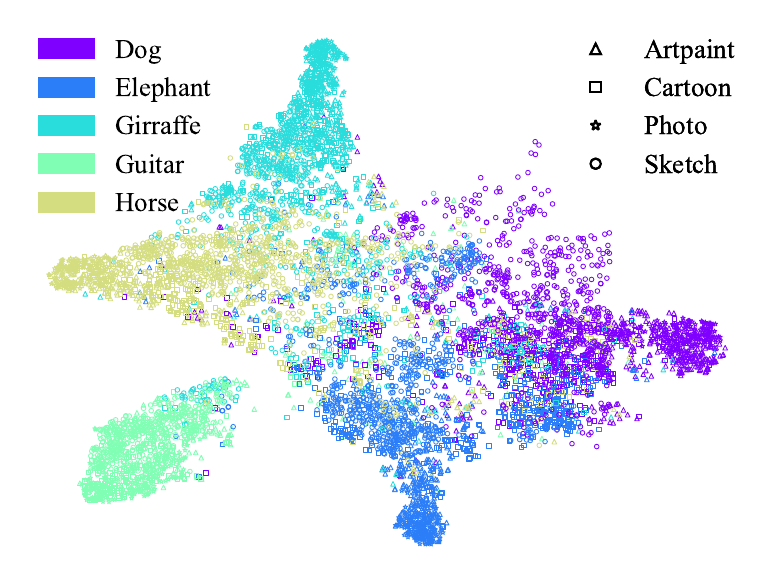} \label{fig:subfig4} } 
         \subfloat[Ours \\ (acc=63.68)]{\includegraphics[width=0.199\linewidth]{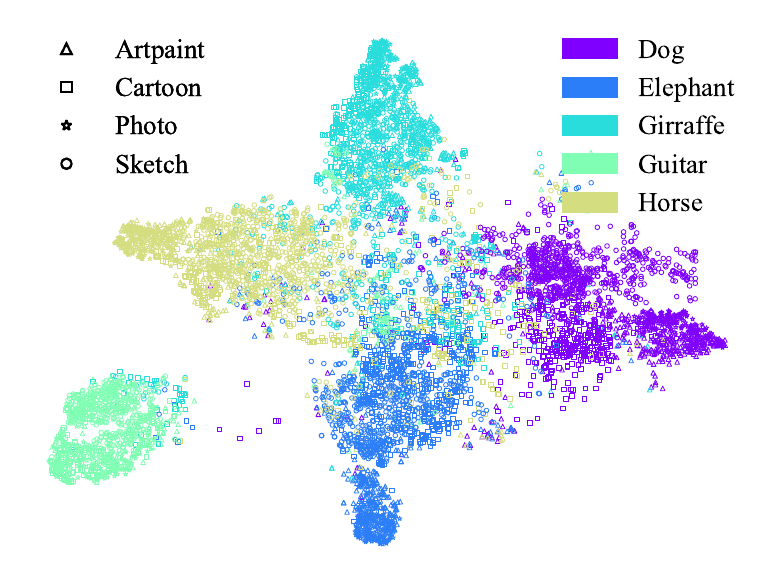} \label{fig:subfig5} }  
      
    \caption{Feature visualization on PACS. $\hollowstar$, $\triangle$, $\Box$, and $\bigcirc$ denote the features of source domain (``Photo'') and other three target domains (``Artpaint'', ``Cartoon'', and ``Sketch''). Different colors denote different categories.}
    \label{feat_visual_different_domains}
\end{figure*}
Furthermore, we also conduct the learned feature visualization shown in Fig.  \ref{feat_visual_different_domains}. From the results, we can see that the learned features for the same category in different domains become more compact, while the ones for different categories become more distinguishable, which suggests that the latent feature augmentation with causal learning and intervention benefits the discriminative power of learned causal features, which is crucial for Single-DG.

\begin{figure}[!t]
  \centering
      \subfloat[Full features]{\includegraphics[width=0.45\linewidth]{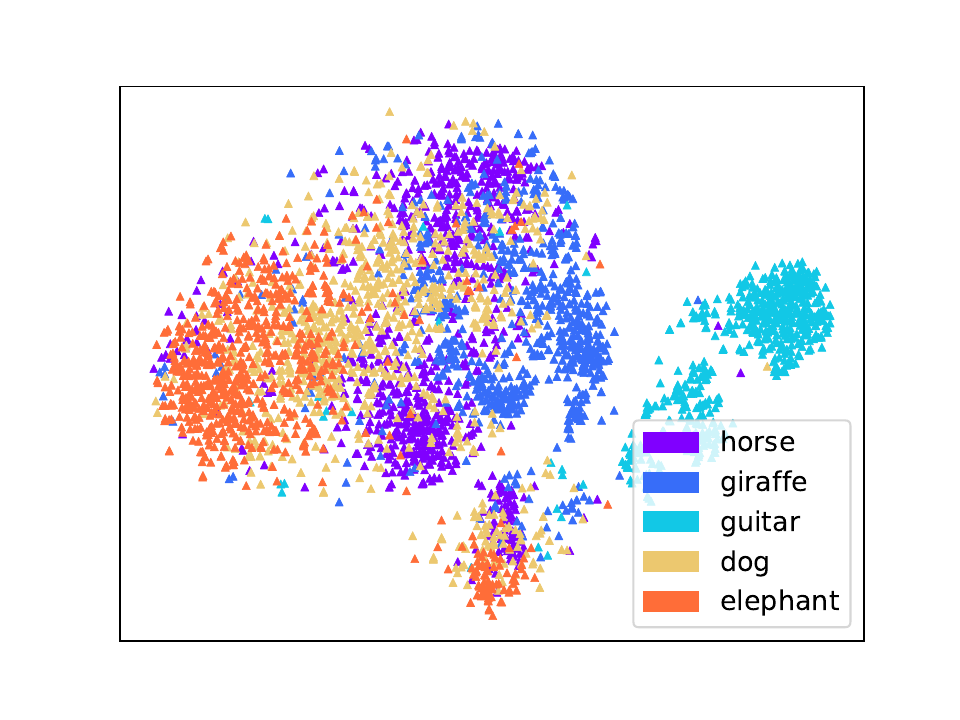} \label{base_feat_vis}}%
       \subfloat[Decoupled features]{\includegraphics[width=0.45\linewidth]{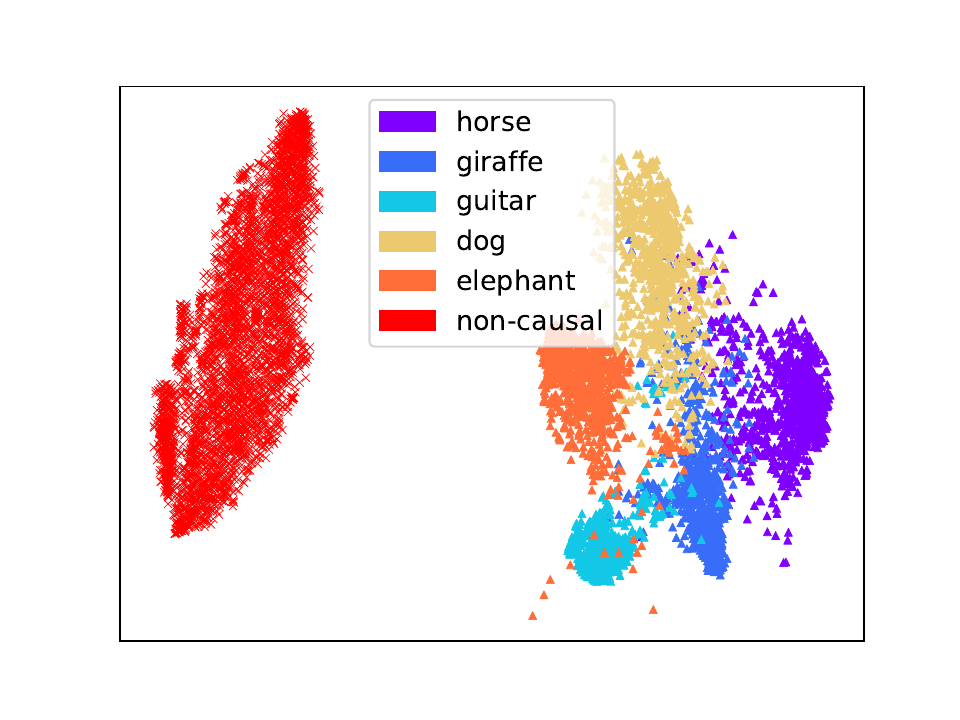} \label{our_causal_noncausal_feat} }%
    \caption{t-SNE visualization of the learned features on target domain ``Sketch'' (source domain: Photo) of PACS dataset. (a) The learned full features of ERM. (b) The decoupled representation of causal and non-causal features by ours. Non-causal features are presented by `$\times$' in red colors, while the causal features are presented by `$\triangle$' in different colors.}
    \label{decoupling_rep_vis}
\end{figure}

Additional, we visualize the learned mixed full features of plain ERM and decoupled features by ours respectively in Fig. \ref{decoupling_rep_vis}. Note that our method effectively disentangles causal and non-causal features, allowing the model to learn better domain-invariant causal features (inter-class samples can be better separated, and intra-class samples can be better clustered) for generalization.

\begin{table}[!t]
  \centering
  \small
  \caption{Ablation study (\%) of causality constraint and intervention module on PACS.}
  \begin{tabular}{@{}l|cccc|c@{}}
    \toprule
    Method &Artpaint & Cartoon & Photo & Sketch & Avg. \\
    \midrule
    baseline    & 70.90  & 76.50 & 42.20 & 53.10 & 60.70 \\
    baseline+$T$ & 74.79  & 77.82 & 56.81 & 59.43 & 67.21 \\ \hline 
    +$\mathcal{L}_{cls}$ & 76.58  & 79.60 & 59.64 & 61.77 & 69.40 \\ 
    +$\mathcal{L}_{cls}$+$\mathcal{L}_{ind}$ & 77.23  & 79.84 & 60.05 & 62.36 & 69.87 \\ 
    +$\mathcal{L}_{cls}$+$\mathcal{L}_{ind}$+$\mathcal{L}_{aug}$ & 79.46  & 80.79 & 62.74 & 62.86 & 71.46 \\  \hline
    Ours & \textbf{81.26}  & \textbf{81.63} & \textbf{63.68} & \textbf{63.44} & \textbf{72.50} \\ 
    \bottomrule
  \end{tabular}
  \label{Ablation_model_causality}
\end{table}

\subsection{Effect of Different Meta-knowledge Learning}
\label{sub:eff_diff_enc}

\begin{figure}[!t]
  \centering
      \subfloat[$E_{ag}$]{\includegraphics[width=0.45\linewidth]{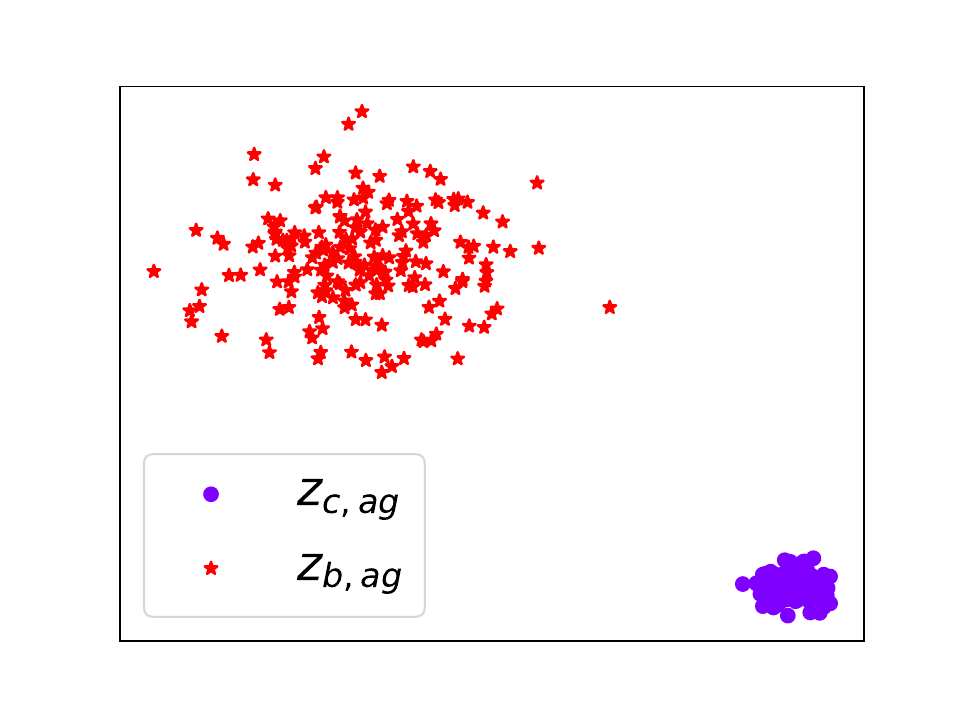} \label{meta_knowledge_visual_E_ag}}%
       \subfloat[$E_{ap}$]{\includegraphics[width=0.45\linewidth]{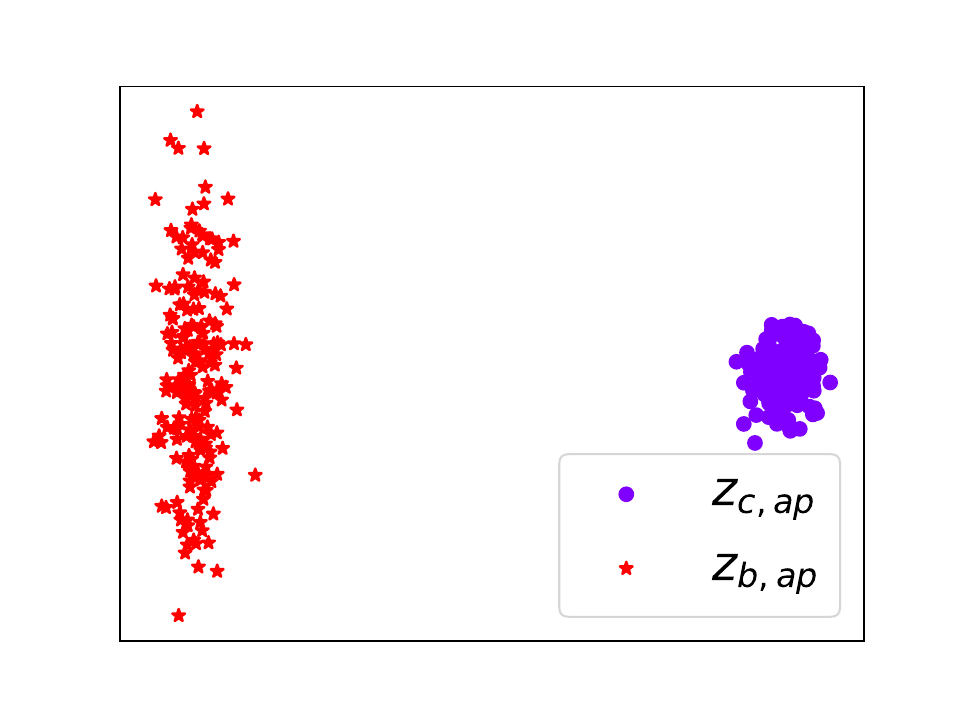} \label{meta_knowledge_visual_E_ap} }%
    \caption{Visualization of learned meta-knowledge about feature-level transformation generated from two encoders. }
    \label{meta_knowledge_visual_E}
\end{figure}

We investigate the roles of the two encoders for different meta-knowledge learning in the latent feature augmentation procedure. As reported in Table \ref{Ablation_model_aug_enc}, both two encoders are capable of learning meta-knowledge, generating a variety of additional implicit feature-level transformation strategies, thereby expanding the source domain distribution and learning better domain-invariant causal features. 
\begin{table}[!t]
  \centering
  \small
  \caption{Ablation study (\%) on the effect of different Encoders on PACS.}
  \begin{tabular}{@{}l|cccc|c@{}}
    \toprule
    Method &Artpaint & Cartoon & Photo & Sketch & Avg. \\
    \midrule
    Ours w/o $E_{ag}$ &79.75  &80.84 &61.88 &63.07 &71.39 \\ 
    Ours w/o $E_{ap}$ &78.92  &80.43 &62.51 &62.86 &71.18 \\ \hline
    Ours & \textbf{81.26}  & \textbf{81.63} & \textbf{63.68} & \textbf{63.44} & \textbf{72.50} \\  
    \bottomrule
  \end{tabular}
  \label{Ablation_model_aug_enc}
\end{table}
Additionally, we visualize the learned meta-knowledge of $E_{ag}$ and $E_{ap}$ respectively in latent feature space with a randomly generated triple $\{x_s^a, x_s^g, x_s^p\}$. We perform $200$ sampling operations from learned Gaussian distribution. As shown in Fig. \ref{meta_knowledge_visual_E}, the meta-knowledge learned by the two encoders for generating feature-level transformation can augment both causal and non-causal features differently. However, it can be observed that the generated implicit feature-level transformation strategies for causal features is more compact, while that for non-causal features vary greatly. This indicates that our approach can generate more diverse implicit feature-level transformations from limited explicit image-level transformations, while simultaneously constraining the stability and invariance of  causal features and the diversity of non-causal features.

\subsection{Sensitive Study}
\subsubsection{Sensitive Study of $\lambda$}
\begin{figure}[!t]
  \centering
      \subfloat{\includegraphics[width=0.8\linewidth]{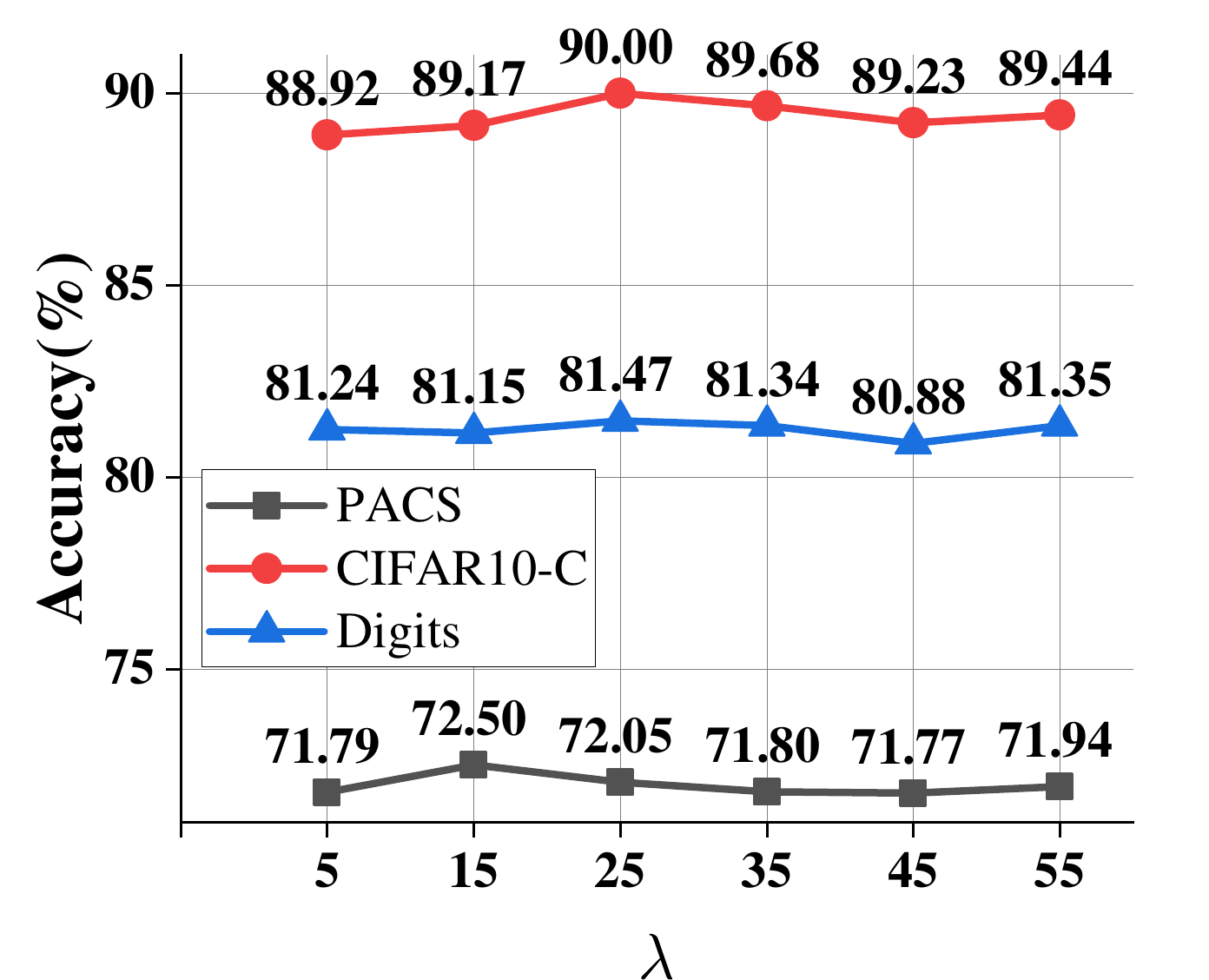}  }%
    \caption{The average accuracy of our method with different values of $\lambda$ on three Single-DG datasets.}
    \label{sens_lambda}
\end{figure}
We investigate the influence of different numbers of sampling operations $\lambda$ in each iteration. As shown in Fig. \ref{sens_lambda}, we select the optimal $\lambda$ value from candidate set $\{5,15,25,35,45,55\}$ in practice. It can be found that the model achieves the best performance on PACS when $\lambda$ is set to $15$. On the CIFAR10-C and Digits datasets, the optimal values for $\lambda$ are both set to $25$ for training. Additionally, the overall performance of our model is relatively stable for other $\lambda$ values.

\subsubsection{Sensitive Study of Loss Coefficients} \label{sens_loss_coeffi_subsec}
\begin{figure*}[!t]
  \centering
    \subfloat{\includegraphics[width=0.32\linewidth]{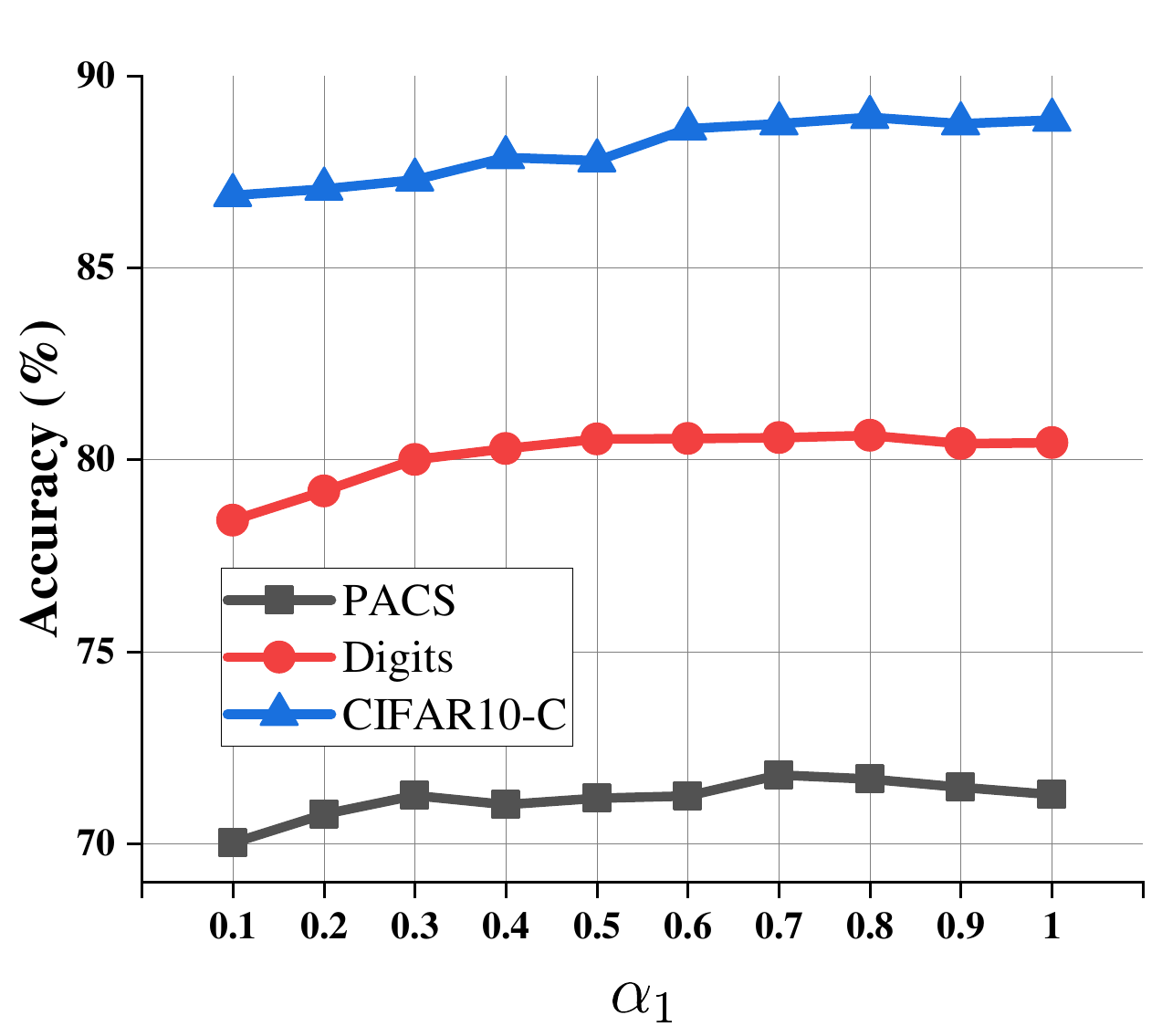} \label{alpha1} }%
      \subfloat{\includegraphics[width=0.32\linewidth]{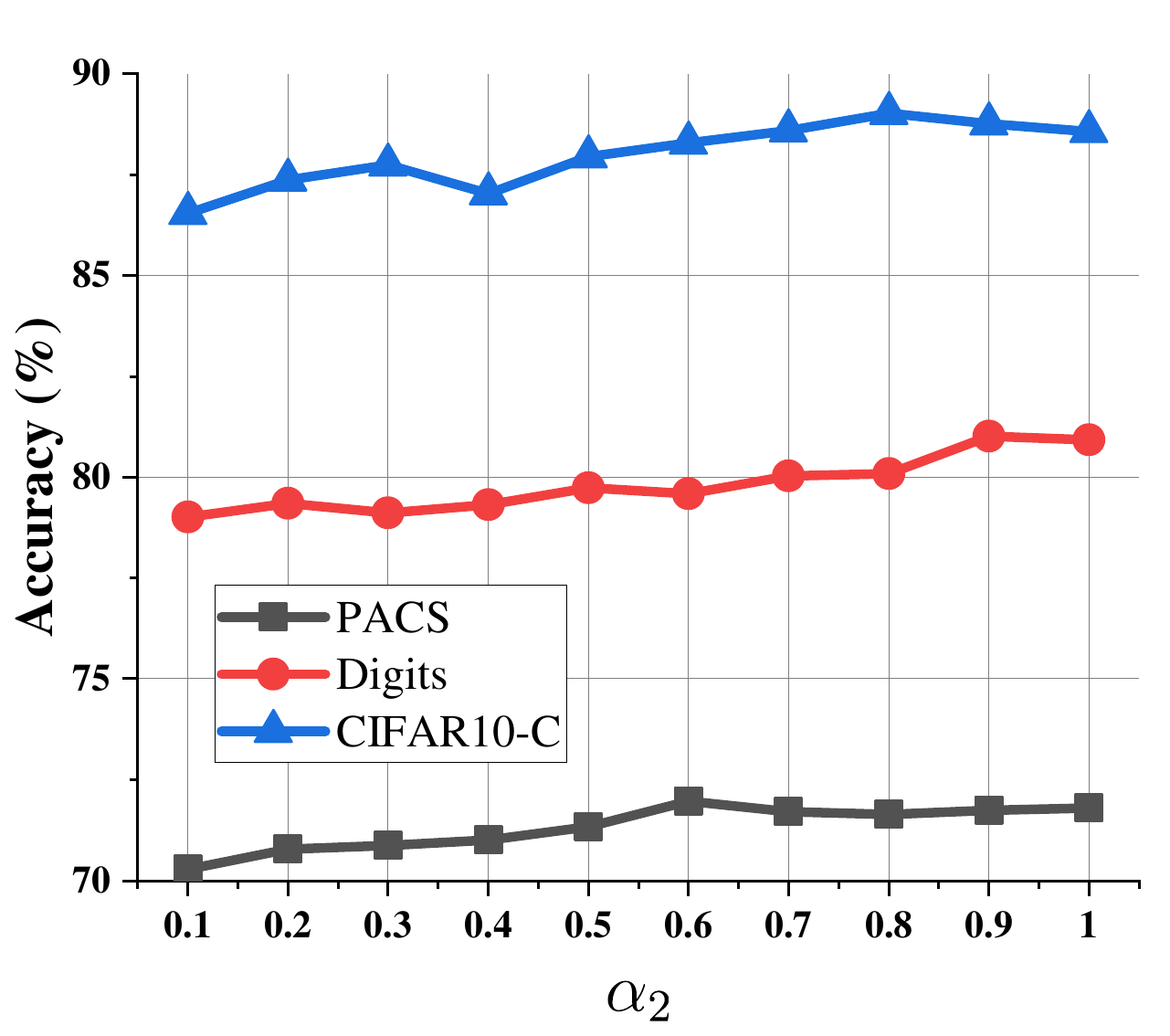} \label{alpha2} }%
    \subfloat{\includegraphics[width=0.32\linewidth]{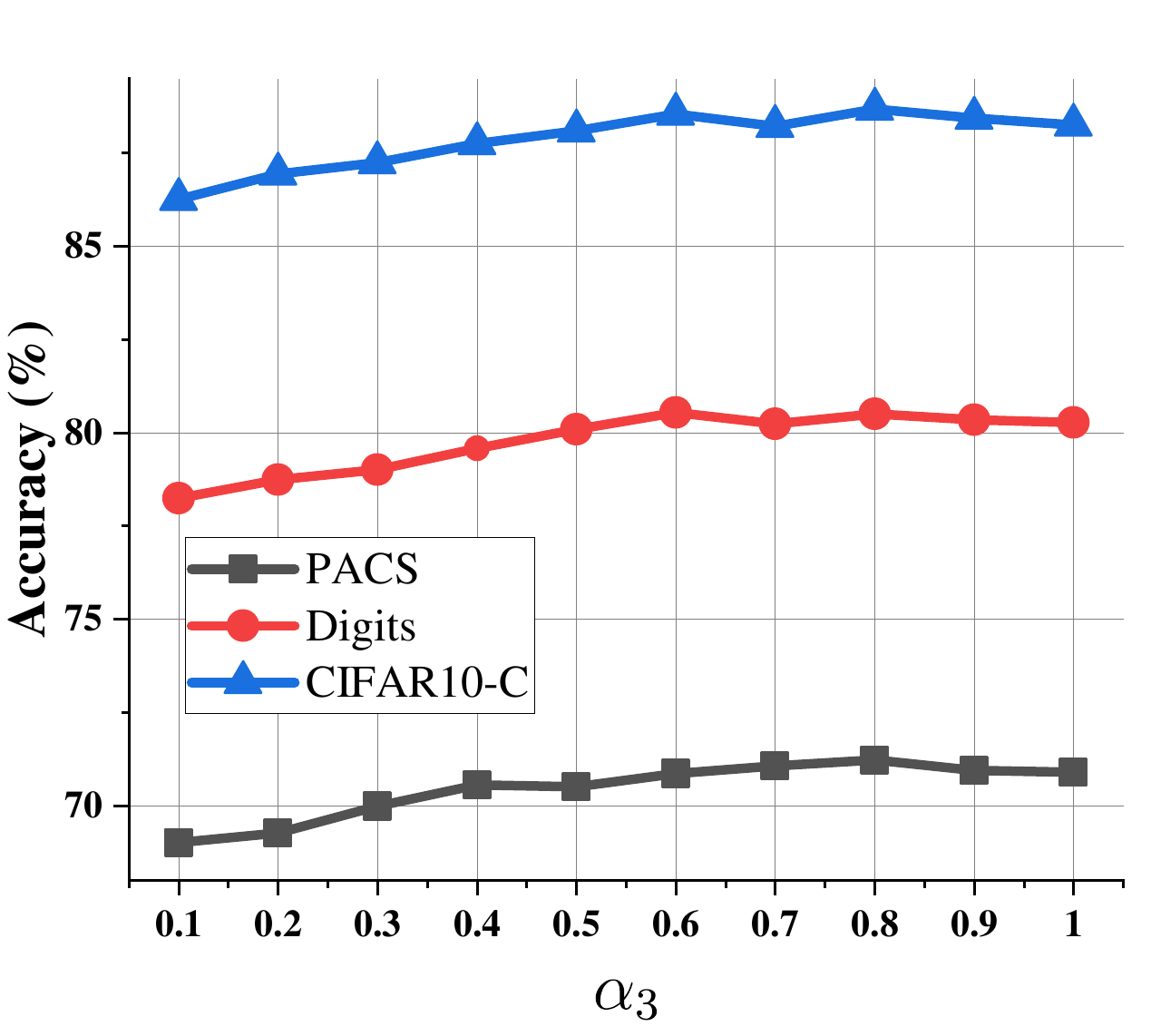} \label{alpha3} }%
    \caption{Sensitivity analysis of three weighting coefficients in Eq. (\ref{total_loss}). }
    \label{loss_param_sens}
\end{figure*}
In Eq. (\ref{total_loss}), $\alpha_1$, $\alpha_2$, and $\alpha_3$ are three weighting coefficients for independence loss, augmentation loss, and causal intervention loss respectively. We simultaneously fix the values of two weight coefficients at $0.5$ while varying the remaining weight coefficient from $0$ to $1$, with a step size of $0.1$. Fig. \ref{loss_param_sens} illustrates how changes in one specific weight affect the loss function while keeping other coefficients constant. From the results, we can see that excessively small weight coefficients lead to a decline in performance. Moreover, once the weight coefficients exceed a certain threshold, our model stabilizes, indicating that our model is not very sensitive to the hyper-parameter settings of weight coefficients when they are relatively large.

\subsection{Analysis of the Quantity and Way of Image-level Transformation Strategy} \label{analy_quan_way_img_level_trans}

\begin{figure*}[!t]
  \centering
      \subfloat[PACS]{\includegraphics[width=0.33\linewidth]{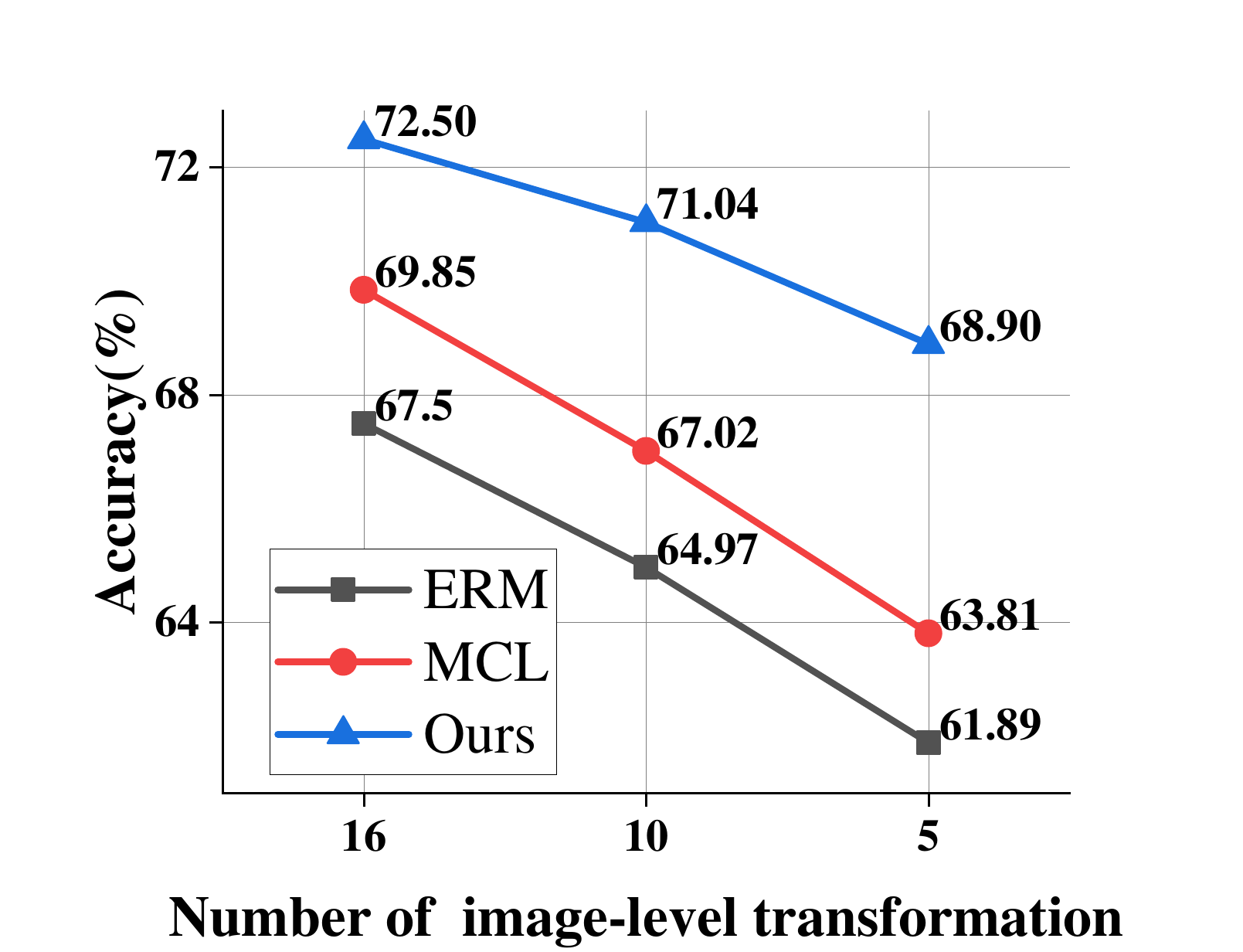} \label{PACS_num_T} }%
    \subfloat[Digits]{\includegraphics[width=0.33\linewidth]{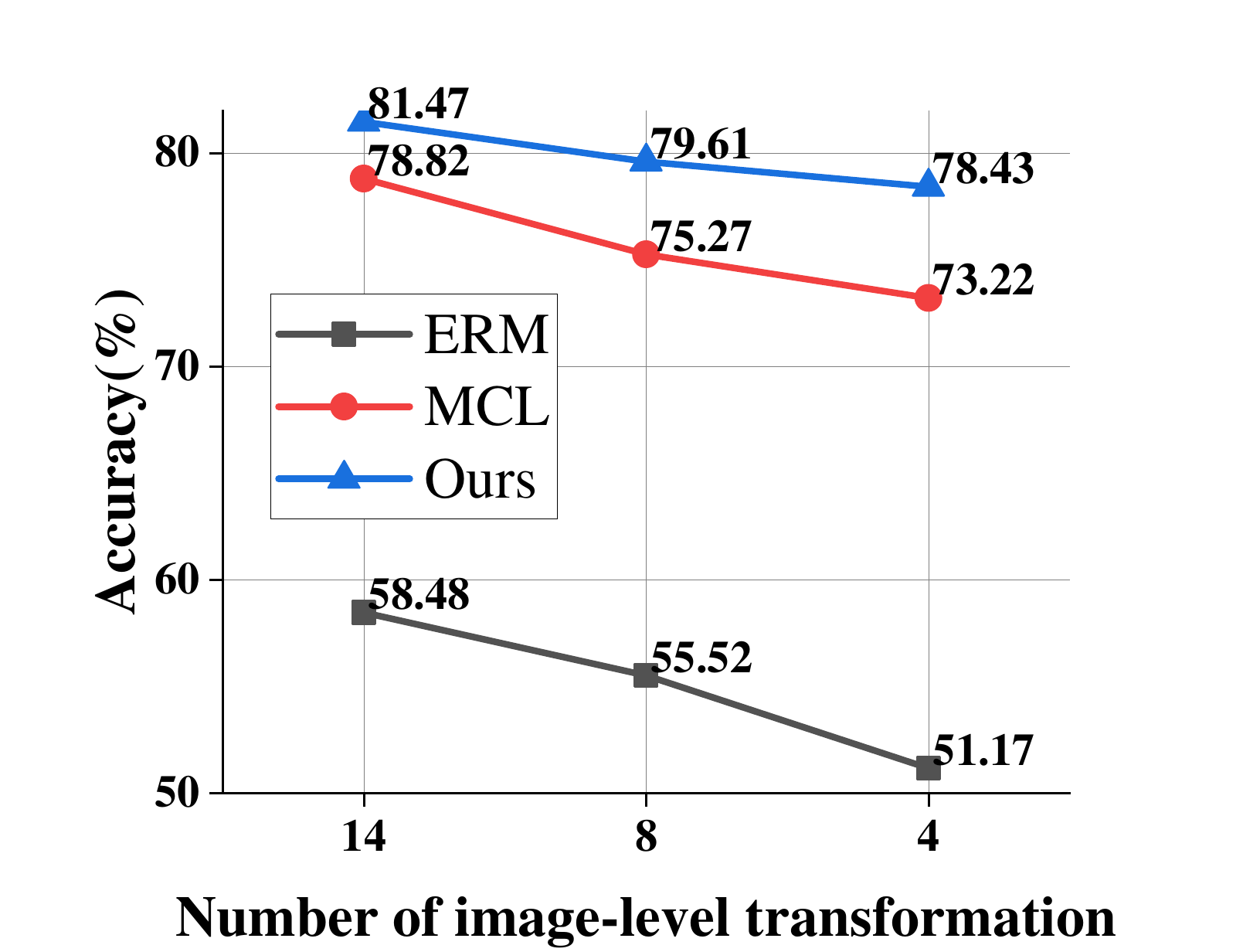} \label{DIGITS_num_T} }%
      \subfloat[CIFAR10-C]{\includegraphics[width=0.33\linewidth]{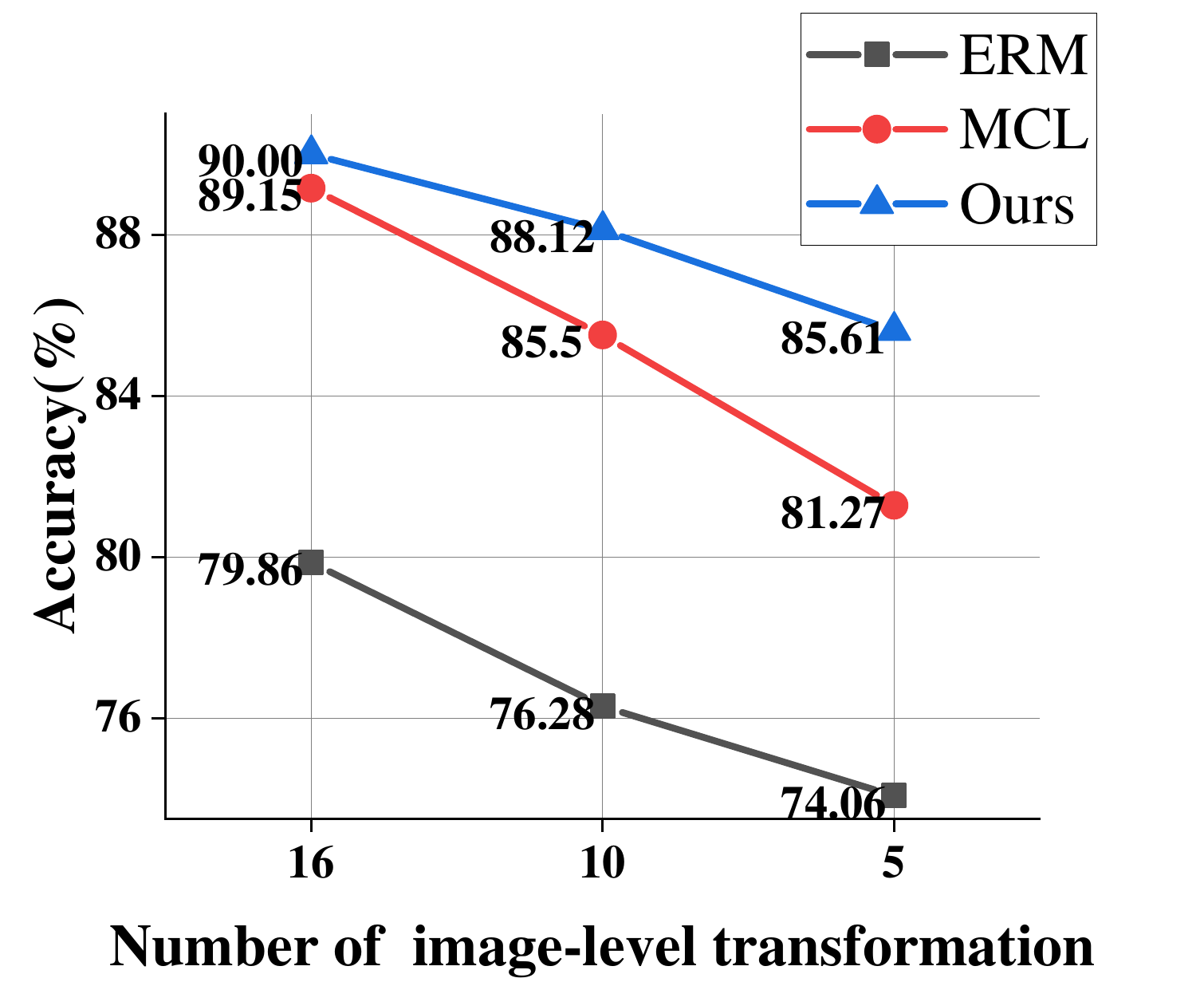} \label{CIFAR_num_T} } 
    \caption{The average accuracy on (a) PACS, (b) CIFAR10-C, and (c) Digits datasets for three models with different quantities of finite image-level transformation strategies. }
    \label{num_T}
\end{figure*}

We conduct a study on the impact of the quantity of finite pre-defined explicit image-level transformation strategies for model performance. We choose ERM~\cite{koltchinskii2011oracle}, MCL~\cite{chen2023meta}, and our method as a comparison. 
Specifically, the quantities of finite pre-defined explicit image-level transformation strategies are set as $\{16,10,5\}$\footnote{\textbf{$16$}: all, \textbf{$10$}: $8$ photometric factors (\textit{Brightness}, \textit{Contrast}, \textit{Color}, \textit{Sharpness}, \textit{AutoContrast}, \textit{Invert}, \textit{Equalize}, \textit{Solarize}) and $2$ geometric factor (\textit{Rotate}, \textit{Flip}), \textbf{$5$}: $4$ photometric factors (\textit{Brightness}, \textit{Contrast}, \textit{Color}, \textit{Sharpness}) and $1$ geometric factor (\textit{Rotate}).} for analysis of PACS and CIFAR10-C datasets and $\{14,8,4\}$\footnote{Since \textit{Rotate} and \textit{Flip} can alter the semantic of digit images, other $14$ strategies are utilized on {Digits}. \textbf{$14$}: all photometric factors and $2$ geometric factors (\textit{Shear-X}, \textit{Shear-Y}), \textbf{$8$}: $6$ photometric factors (\textit{Brightness}, \textit{Contrast}, \textit{Color}, \textit{Sharpness}, \textit{AutoContrast}, \textit{Invert}) and $2$ geometric factors (\textit{Shear-X}, \textit{Shear-Y}), \textbf{$4$}: $2$ photometric factors (\textit{Brightness}, \textit{Contrast}, \textit{Color}, \textit{Sharpness}) and $2$ geometric factors (\textit{Shear-X}, \textit{Shear-Y}).} for analysis for Digits dataset.
From Fig. \ref{num_T}, with the reduction in the number of image-level transformation strategies, all methods exhibit a decrease in generalization performance. However, our method displays better stability and robustness, which profits from our latent feature augmentation and generates infinite implicit feature-level transformation strategies based on the two types of meta-knowledge. Therefore, we can reduce the dependence on initial finite image-level transformation strategies and still learn stable domain-invariant causal features.

\begin{figure}[!t]
  \centering
\includegraphics[width=0.7\linewidth]{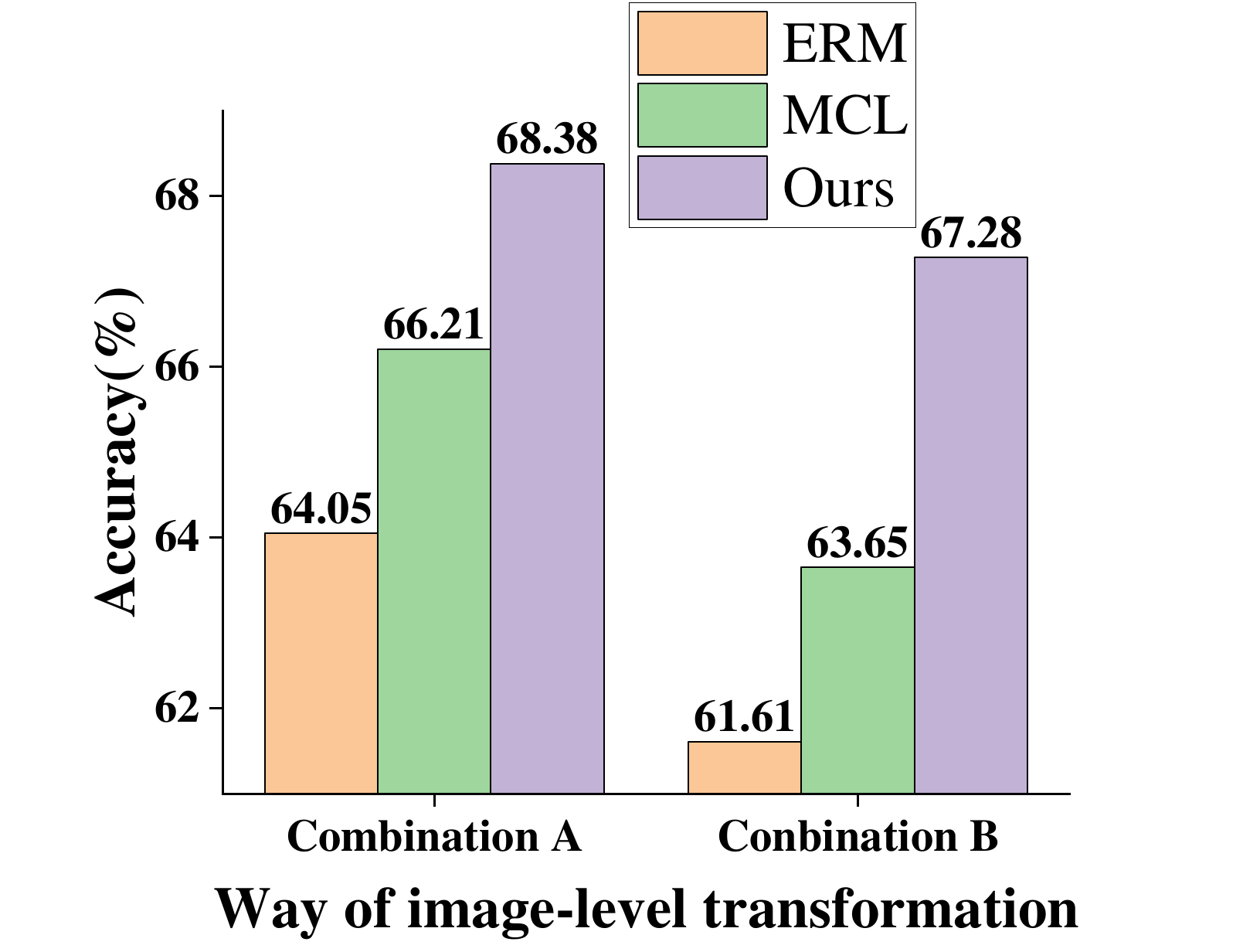}    
    \caption{The average accuracy on PACS for three models with different image-level ways of transformation strategies combination. }
    \label{acc_diff_aug}  
\end{figure}

Besides, an additional experiment is conducted to validate the effectiveness of our method under image-level transformation strategies with different diversity. Specifically, we design two strategies combination, i.e., \textbf{Combination A}: \textit{Rotate+NoiseSalt+Solarize} and \textbf{Combination B}: \textit{Rotate+NoiseSalt+NoiseGaussian}. Obviously, Combination A has richer diversity than B. The result is shown in Fig. \ref{acc_diff_aug}, when we use a combination strategy with fewer changes, the performance of all models exhibits varying degrees of decline. This suggests that the model's generalization ability is related to the specific image-level transformation strategies and the transformation strategies lacking diversity are not sufficient to simulate the unseen domain which results in relatively poor performances. Moreover, our performance using the transformation Combination B is even higher than that of other methods using Combination A. This shows that our latent feature augmentation can effectively compensate for the lack of diversity in image-level transformation strategies. 

\begin{table}[!t]
  \centering
  \small
  \caption{Comparison of computational cost along with the performance on PACS during inference. Tested with the image size of $227\times 227$ on one Quadro RTX 8000 GPU. We average the inference time over $1000$ trials. (GMACs: Giga Multiply-Accumulate Operations)}
  \begin{tabular}{@{}l|cccc@{}}
    \toprule
    Method & \# of Params & GMACs & Time (ms) & Acc.(\%)\\
    \midrule
    ERM~\cite{koltchinskii2011oracle}    & 11.191M  & 2.062 & 1.892 &60.70 \\
    PDEN~\cite{Li_Gao_Cao_Huang_Weng_Mi_Yu_Li_Xia_2021} & 11.191M  & 2.062 & 1.917 &65.68 \\
    L2D~\cite{Wang_Luo_Qiu_Huang_Baktashmotlagh_2021}    & 19.584M  & 2.070 &  2.153  &65.41\\
    MCL~\cite{chen2023meta}    & 17.488M  & 35.149 & 27.304 &69.86 \\
    Ours    & 13.282M  & 2.064 & 1.976 &72.50 \\
    \bottomrule
  \end{tabular}
  \label{complexity_net}
\end{table}

\subsection{Example Visualization} \label{example_visual_sub_sec}
\begin{figure*}[!t]
  \centering 
      \subfloat{\includegraphics[width=0.8\linewidth]{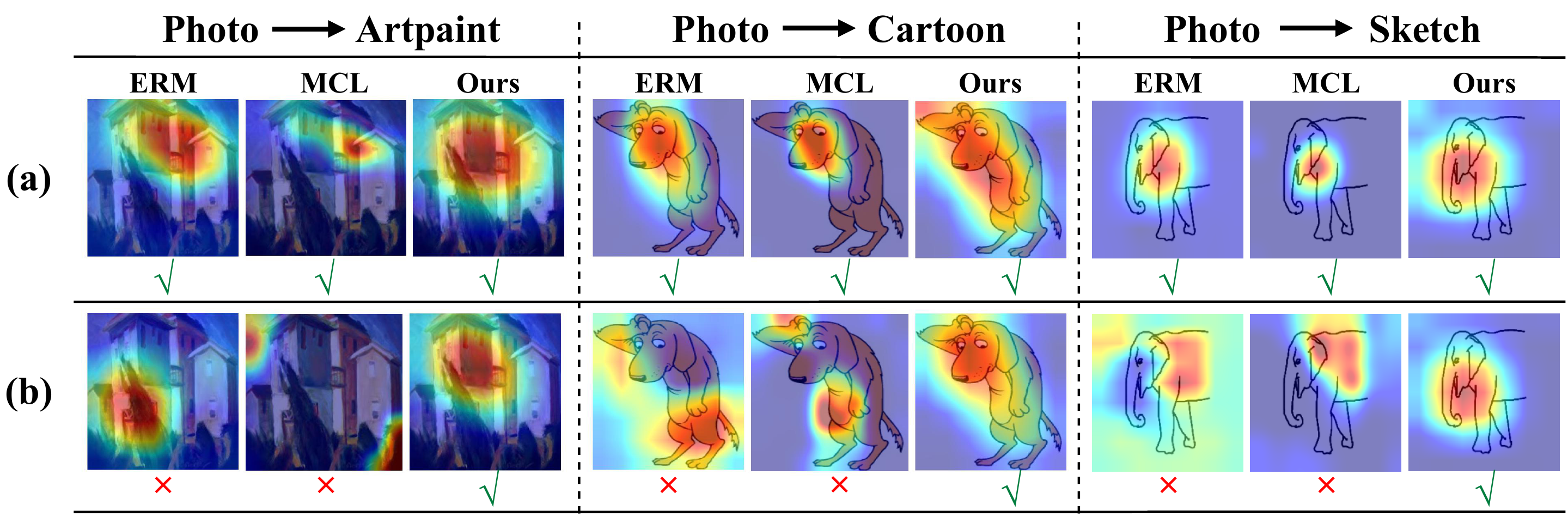} }%
    \caption{Example visualization for three models ERM, MCL, and Ours on Single-DG with (a) $16$ or (b) $5$ types of image-level transformations. ($\color{green}{\checkmark}$: correct prediction, $\color{red}{\times}$: wrong prediction)  } 
    \label{different_aug_for_different_causal}
\end{figure*}

In Fig. \ref{different_aug_for_different_causal}, we conduct the visualization of class activation maps  for ERM~\cite{koltchinskii2011oracle}, MCL~\cite{chen2023meta}, and Ours on Single-DG task. First, we find that when the quantity of image-level transformation strategy is reduced from $16$ to $5$, ERM and MCL focus on non-causal factors which leads to misclassification, while our method can still pay attention to causal factors. This indicates that our method can learn more stable and robust domain-invariant causal features and does not exhibit excessive dependence on the quantity of initial transformation strategies in domain-invariant causal feature learning. Second, as shown in the examples on the ``Sketch'' domain, although ERM and MCL using $16$ image-level transformation strategies can make correct predictions, the regions of interest are not always causal features. In contrast, our method focuses on regions of the image that correspond to causal parts and makes correct predictions. This indicates that our method excels in domain-invariant causal feature learning, which is crucial for Single-DG.

\begin{table*}[!t]
  \centering
  \scriptsize
  \caption{Leave-one-domain-out results on PACS, Office-Home, and DomainNet benchmark with ResNet50 ImageNet pre-trained model (Average accuracy(\%)). MCL$^\dag$ denotes our rerun of MCL using the authors' official code following the same settings in the original paper~\cite{chen2023meta}. }
  \begin{tabular}{@{}l|cccc|c||cccc|c||cccccc|c@{}}
    \toprule
    \multirow{2}{*}{Method} &\multicolumn{5}{c}{PACS} &\multicolumn{5}{c}{Office-Home} &\multicolumn{7}{c}{DomainNet} \\ \cline{2-18}
    & A &C &P &S &Avg. &Ar &Cl &Pr &Rw &Avg. & clip &info &paint &quick &real &sketch &Avg. \\  
    \midrule
    IRM~\cite{arjovsky2019invariant} &84.8 &76.4 &96.7 & 76.1 &83.5 &58.9 &52.2 &72.1 &74.0  &64.3 &48.5 &15.0 &38.3 &10.9 &48.2 &42.3 &33.9\\
    DANN~\cite{ganin2016domain} &86.4 &77.4 &97.3 &73.5 &83.6 &59.9 &53.0 &73.6 &76.9  &65.9 &53.1  &18.3  &44.2  &11.8 &55.5 &46.8  &38.3\\ 
    Mixup~\cite{xu2020adversarial} &86.1 &78.9 &97.6 &75.8 &84.6 &62.4 &54.8 &76.9 &78.3  &68.1 &55.7 &18.5 &44.3 &12.5 &55.8 &48.2 &39.2\\ 
    MMD~\cite{li2018domain} &86.1 &79.4 &96.6 &76.5  &84.6  &60.4 &53.3 &74.3 &77.4  &66.3  &32.1 &11.0 &26.8 &8.7 &32.7 &28.9  &23.4 \\ 
    MLDG~\cite{li2018learning} &85.5 &80.1 &97.4  &76.6  &84.9 &61.5 &53.2 &75.0 &77.5  &66.8 &59.1 &19.1 &45.8 &13.4 &59.6 &50.2 &41.2 \\ 
    RSC~\cite{huang2020self} &85.4 &79.7 &97.6  &78.2 &85.2 &60.7 &51.4 &74.8 &75.1 &65.5 &55.0 &18.3 &44.4 &12.2 &55.7 &47.8 &38.9 \\ 
    SagNet~\cite{nam2021reducing} &87.4 &80.7 &97.1 &80.0  &86.3 &63.4 &54.8 &75.8 &78.3 &68.1 &57.7 &19.0 &45.3 &12.7 &58.1 &48.8 &40.3 \\ 
    CORAL~\cite{sun2016deep} &88.3 & 80.0  &97.5 &78.8 &86.2 &65.3 &54.4 &76.5 &78.4 &68.7 &59.2 &19.7 &46.6 &13.4 &59.8 &50.1 &41.5 \\  
    iDAG~\cite{huang2023idag} &\textbf{90.8} &83.7 &9\textbf{8.0} &82.7 &88.8 &68.2 &57.9 &79.7 &81.4 &71.8 &67.9 &24.2 &55.0 &16.4 &66.1 &\textbf{56.9} &47.7 \\   
    MCL$^\dag$~\cite{chen2023meta} &90.6 &85.1 &97.6 &88.6 &90.5 &67.6 &56.3 &\textbf{80.1} & 82.0 &71.5 &68.1 & 25.4 &54.7 & 18.6 & 65.7 & 56.8 &48.2 \\ \hline
    Ours &90.6 &\textbf{86.3} &97.8 &\textbf{89.7}  &\textbf{91.1}  & \textbf{69.6} &\textbf{58.7} &79.9 &\textbf{82.6}  &\textbf{72.7} &\textbf{69.4} &\textbf{27.6} &\textbf{56.3} &\textbf{19.5} &\textbf{67.9} &56.4 &\textbf{49.5} \\
    \bottomrule
  \end{tabular}
  \label{multi_dg_detailed_res}
\end{table*}

\subsection{Computational Efficiency Analysis}
As reported in Table \ref{complexity_net}, the computational efficiency of various Single-DG methods is summarized. Regarding memory and time consumption, since we need to map the initial features to causal features during inference, the number of parameters and computational cost in our method is slightly higher than the baseline method. In sum, the parameter scale and time consumption of our model are comparable with many other state-of-the-art methods. In particular, even though the latest Single-DG method, MCL~\cite{chen2023meta}, achieves the second-best performance, its computational cost is exceptionally high. This is because MCL still requires image-level transformations and analyzes causal effects during inference to achieve causal feature mapping. In contrast, our method not only achieves the current best generalization performance but also uses only one well-learned linear layer for causal feature mapping during inference.

\subsection{Results on Multiple Source Domain Generalization}

We extend our method to Multi-DG setting without using domain labels by employing the leave-one-domain-out protocol following existing Multi-DG experimental setting~\cite{gulrajani2020search}. PACS and Office-Home contain  diversity shift while the DomainNet is a more realistic and challenging task which contains both diversity and correlation shift. The average results are shown in Table \ref{multi_dg_detailed_res}, which demonstrates that our method is also capable of handling multi-source domain settings and get the best generalization performance over the existing Multi-DG methods. Particularly, our method outperforms current methods that incorporate causal inference, such as iDAG~\cite{huang2023idag} and MCL~\cite{chen2023meta}.

\section{Conclusion}

In this work, we present a novel latent feature augmentation paradigm leveraging augmentations on the latent representation of instances, which significantly improves the diversity of observable domain shifts. An effective intervention manner is designed to expand the diversity of distributions and learn more stable causal features for generalization. Our proposed method can reduce the strong reliance on finite image-level transformation. Extensive experiments demonstrated the effectiveness of our work for Single-DG and also Multi-DG.

\bibliography{ref}
\bibliographystyle{IEEEtran}

\newpage

\vfill

\end{document}